# Multi-Resolution SAR and Optical Remote Sensing Image Registration Methods: A Review, Datasets, and Future Perspectives

**Wenfei Zhang, Ruipeng Zhao, Yongxiang Yao[1,2*], Yi Wan[1,2], Peihao Wu[1], Jiayuan Li[1], Yansheng Li[1], Yongjun Zhang[1,2*]**

[1]School of Remote Sensing and Information Engineering, Wuhan University, Wuhan 430079, China;
[2]Technology Innovation Center for Collaborative Applications of Natural Resources Data in GBA, Ministry of Natural Resources, No.468 Huanshi East Road, Yuexiu District, Guangzhou, Guangdong Province, 510075, China
zhangwenfei@whu.edu.cn (W.Z.); zhaorp@whu.edu.cn(R.Z.); yaoyongxiang@whu.edu.cn (Y.Y.); yi.wan@whu.edu.cn (Y.W.); wupeihao@whu.edu.cn (P.W.) ; ljy_whu_2012@whu.edu.cn(J.L.); yansheng.li@whu.edu.cn(Y.L.) ; zhangyj@whu.edu.cn (Y.Z.);
* Corresponding author: zhangyj@whu.edu.cn and yaoyongxiang@whu.edu.cn

**Abstract**

Synthetic Aperture Radar (SAR) and optical image registration is crucial for multi-source remote sensing data fusion, with widespread applications in military reconnaissance, environmental monitoring, and disaster management. However, significant differences in imaging mechanisms, geometric distortions, and radiometric properties between SAR and optical images pose persistent challenges. As image resolution improves, fine SAR textures become more critical, leading to increased overlap, misalignment, and three-dimensional spatial discrepancies. Meanwhile, two major research gaps exist: the lack of a publicly available multi-resolution, multi-scene registration dataset and the absence of a systematic, multi-level analysis of current registration methods. In response, this study summarizes registration research progress from the perspective of data resolution, providing detailed insights into challenges and future trends. To further advance method evaluation, the MultiResSAR dataset was constructed and released, comprising over 10k pairs of multi-source, multi-resolution, and multi-scene SAR and optical images. Sixteen state-of-the-art algorithms were tested. The results show that: (1) No algorithm achieves 100% matching success across varying resolutions and scenes, and performance deteriorates as resolution increases, with almost all failing on sub-meter data; (2) Among deep learning approaches, XoFTR performs best with a 40.58% success rate, while among traditional methods, RIFT achieves the highest success rate at 66.51%. Most other algorithms fall below 50%; (3) Future research must address noise suppression, fusion of three-dimensional geometric information, cross-view geometric transformation modeling, and deep learning model optimization to achieve robust registration of high-resolution SAR and optical images. The source dataset in this work is available at https://github.com/betterlll/Multi-Resolution-SAR-dataset- .

## 1.Introduction

As the global demand for Earth observation technology continues to rise, the collaborative processing of multi-source data has gained increasing attention. In this context, SAR (Synthetic Aperture Radar), as one of the key technologies, has found widespread application due to its unique advantage of providing high-resolution surface information under adverse weather conditions. The fundamental principle of SAR is to emit microwave electromagnetic waves toward the ground and receive the echo signals reflected from the surface, thereby acquiring detailed information about the Earth's surface (Tomiyasu, 1978). Since microwave signals can penetrate clouds, rain, and fog, SAR operates in all weather and at all times, providing a solid foundation for multi-source data fusion. With the continuous increase in the number of satellites, high-resolution satellite data has become increasingly abundant, leading to Earth observation data characterized by multiple viewing angles and high revisit rates. This has accelerated the development of multi-sensor remote sensing technologies and greatly promoted the application of SAR data, including areas such as land cover change monitoring, vegetation change tracking, glacier dynamics observation, real-time assessment and emergency response to natural disasters like earthquakes, floods, and landslides, as well as resource development, environmental protection, battlefield surveillance, target reconnaissance, and military intelligence collection.

The development of SAR sensors has undergone multiple stages, from concept verification to meeting the demands of high-resolution observation. In the 1950s, the concept of SAR was first proposed and verified. In the 1980s, the Seasat satellite, equipped with a SAR sensor, achieved the first spaceborne application of SAR. Since the beginning of the 21st century, SAR technology has continuously evolved, covering multiple frequency bands (such as L-band and X-band) and full-polarization techniques, significantly enhancing capabilities in land cover classification and target recognition. In recent years, the resolution of SAR images has continuously improved, from medium-to-low resolutions of around 10 meters to high resolutions of 2–3 meters and even sub-meter resolutions, greatly enhancing the ability to capture fine details. For example, the European Sentinel-1 satellite provides 10-meter resolution images, while China's GF-3 satellite offers C-band multi-polarization SAR images with a resolution of 1 meter, further improving the accuracy of land feature monitoring. Wuhan University's Luojia-2 01 satellite provides Ka-band SAR images with a resolution of 0.5 meters, meeting the demands of detailed observation. Additionally, high-resolution SAR images with resolutions of 25 centimeters and 16 centimeters, respectively, have been released by ICEYE and Umbra, greatly enhancing the ability to detect small objects, such as ships and military equipment.

Despite the continuous development of SAR technology, many challenges remain in the registration process with optical remote sensing images. The presence of geometric deformations, differences in radiometric characteristics, and variations in sensor parameters between SAR and optical images make high-precision registration complex and difficult. First, in terms of geometric deformation, both SAR and optical images are influenced by factors such as terrain variations and atmospheric distortions, resulting in nonlinear transformation relationships. This requires the use of nonlinear registration methods for processing. Second, regarding radiometric characteristics, SAR images are formed through radar reflection signals and possess polarization information and unique radiometric properties, while optical images are affected by optical scattering and transmission. Thus, the radiometric differences between the two types of images need to be addressed through radiometric correction methods. Finally, regarding sensor parameters, the differences in wavelength, polarization modes, and observation angles between SAR and optical sensors increase the difficulty of registration. Thanks to the relentless efforts of researchers, the registration problem between medium- and low-resolution SAR and optical images has been solved to varying degrees. However, with the development of high-resolution images, new challenges have emerged. The detailed texture and three-dimensional structural information in high-resolution SAR images can no longer be ignored, gradually becoming the primary influencing factor in the registration process, as shown in Fig. 1. Therefore, overcoming the impact of complex three-dimensional structures, such as buildings, on image registration has become a key issue for high-precision registration of high-resolution SAR and optical images, holding significant scientific research importance.

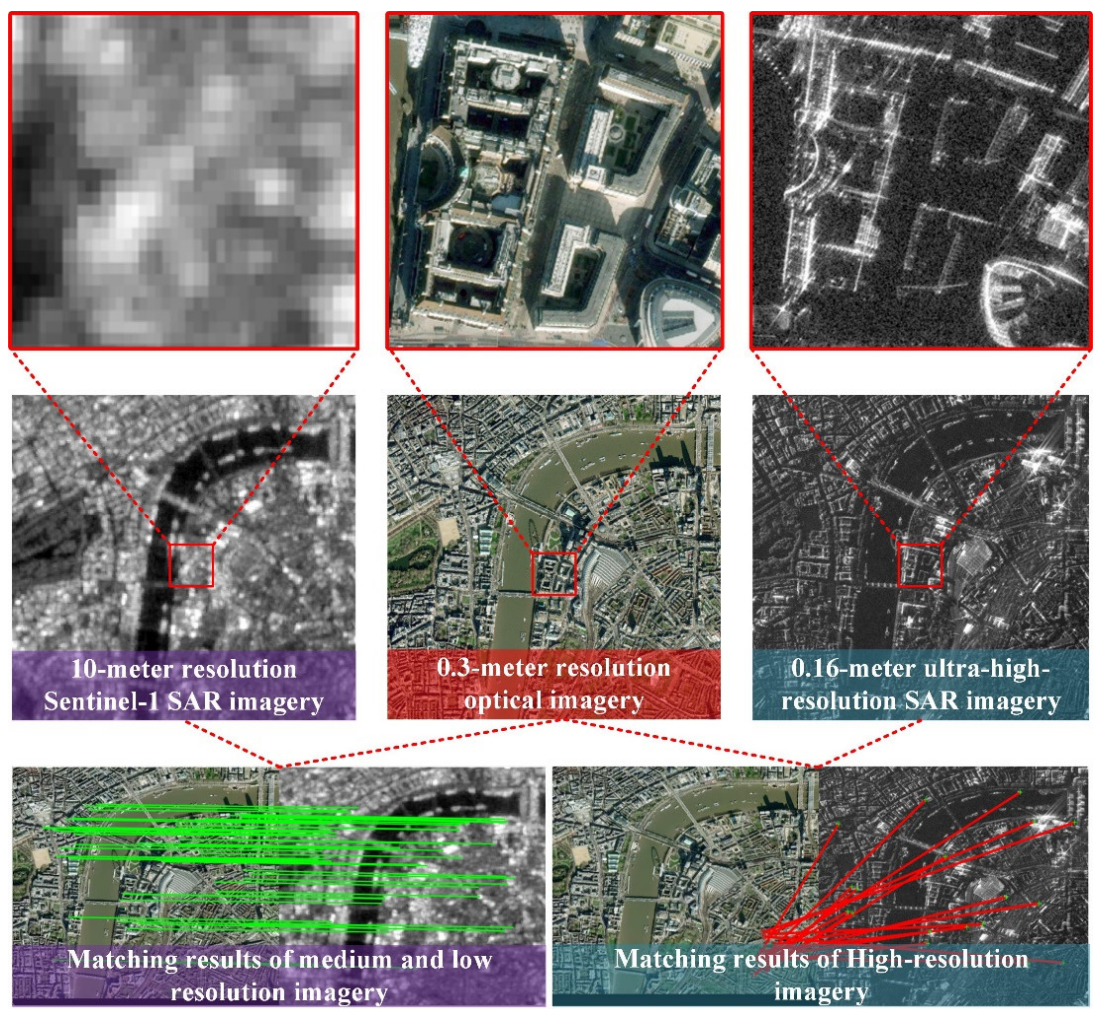

Fig. 1 SAR and Optical Remote Sensing Images at Different Resolutions

In recent years, registration techniques have evolved from initial area-based algorithms to feature-based algorithms, such as SIFT (Scale-Invariant Feature Transform), SURF (Speeded-Up Robust Features), and ORB (Oriented FAST and Rotated BRIEF). These algorithms can extract distinctive and stable feature points from images, enabling accurate matching between them. With the development of deep learning techniques, an increasing number of studies have begun to apply deep learning methods to the problem of SAR and optical remote sensing image registration. Deep learning technology has advanced from the extraction of local features to global feature extraction and eventually to end-to-end deep networks, significantly improving the accuracy and robustness of registration. For example, CNN (Convolutional Neural Network)-based registration methods can directly learn the nonlinear transformation relationships between images, avoiding the complex feature extraction and matching processes inherent in traditional methods, and yielding better results. Some recent review papers have categorized and summarized image registration methods; however, a comprehensive analysis of SAR and optical remote sensing image registration theories and methods is still insufficient. Particularly, with the proliferation of high-resolution SAR images, SAR and optical remote sensing image registration faces even greater challenges, as high-resolution images contain more complex details and information, larger data volumes, more sensitive geometric changes, and more complex radiometric characteristics. This requires registration algorithms to more accurately capture and process these variations in order to ensure the accuracy and stability of the registration.

Overall, although significant progress has been made in various methods and applications, there is still a lack of a systematic review and summary of the research progress on SAR and optical image registration at different resolutions. Additionally, there is a lack of a comprehensive validation dataset to objectively evaluate the performance of existing registration algorithms and the challenges they face. This paper, from the perspective of data resolution, aims to systematically summarize the challenges and current status of SAR and optical remote sensing image registration, propose strategies to address new challenges, and explore future development directions to promote the implementation of this technology in a wide range of applications, thereby improving the accuracy of land feature extraction and analysis. The overall organization is presented in Fig. 2.

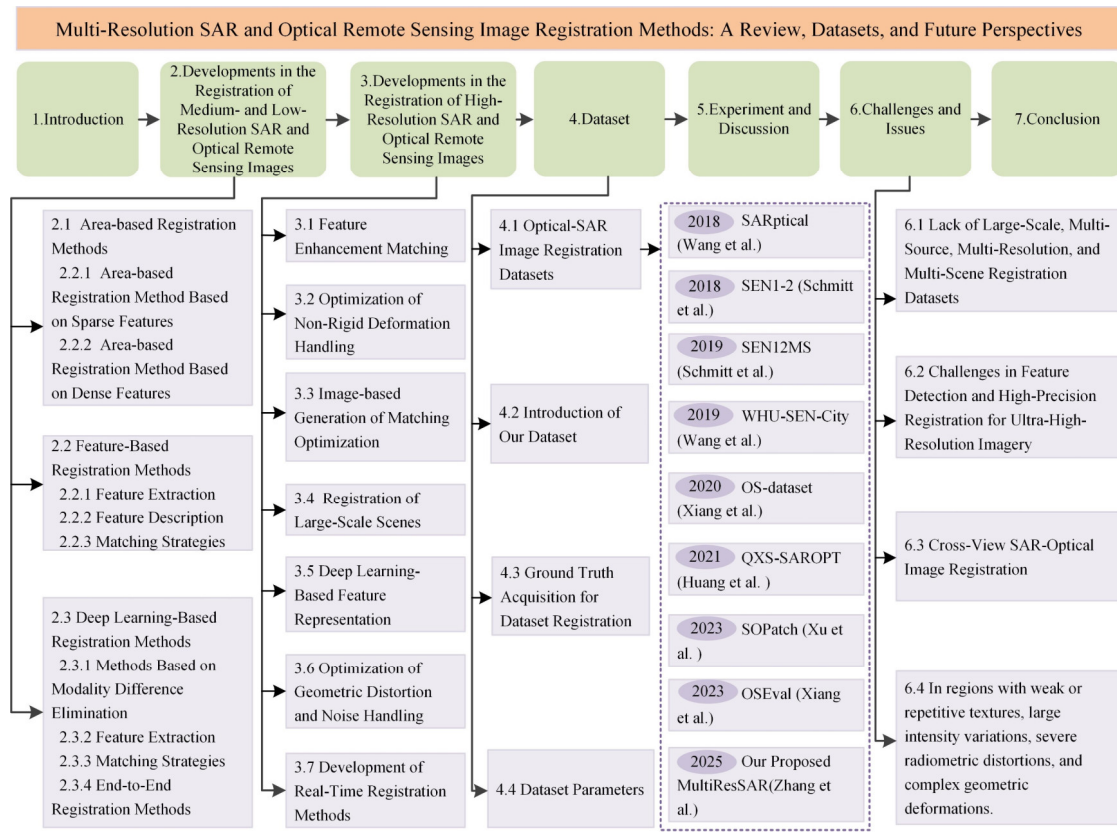

Fig. 2 Structure of this survey

## 2. Developments in the Registration of Medium- and Low-Resolution SAR and Optical Remote Sensing Images

This paper reviews the current development of medium- and low-resolution SAR and optical remote sensing image registration methods, categorized into Area-based methods, feature-based methods, and deep learning-based methods. It primarily focuses on the SAR and optical remote sensing image registration methods proposed in recent years.

## 2.1 Area-based Registration Methods

The core of area-based registration methods lies in selecting an appropriate similarity metric and using it in conjunction with optimization techniques to accurately estimate the geometric transformation parameters between images, thereby driving the optimization of the entire registration process through a template matching strategy (Dowman,1998). For this reason, these methods are also referred to as template-based registration methods. In such methods, similar regions are typically referred to as homologous regions, and their central points are known as homologous points.

In area-based registration algorithms, commonly used similarity metrics include SSD (Sum of Squared Differences), NCC (Normalized Cross-Correlation), MI (Mutual Information), PC (Phase Correlation), and LSS (Local Self-Similarity), among others. SSD is simple to compute and easy to implement, but it is sensitive to lighting changes and noise. NCC is insensitive to lighting variations and brightness differences, making it suitable for handling scale and translation changes, but it requires a large computational load and lacks robustness to rotation and viewpoint changes. MI is effective for registering images with different grayscale but similar content, but it has high computational complexity and is sensitive to large-scale transformations. PC is robust to translation and partial rotation changes, but it performs poorly with more complex non-translation transformations. LSS performs well in handling textured images or images with locally repetitive structures but has high computational complexity and limited robustness to large-scale transformations. In SAR and optical image registration, due to their differing characteristics and transformations, NCC is often considered a more suitable choice, while LSS can be used as a supplementary method to account for local structures and texture information.

Area-based image registration methods can further be divided into sparse registration and dense registration based on the density of the extracted features. The core of sparse feature registration methods is to match significant points or feature points extracted from the image. A major advantage of such methods is that they can effectively reduce computational load and improve matching efficiency, making them particularly suitable for scenarios where there are noticeable structural differences in the image. In contrast, dense feature-based region registration methods focus on analyzing the features of each pixel or a group of adjacent pixels in the image. By calculating and registering the features of various local regions across the entire image, these methods provide a more comprehensive description of the overall structure of the image. Dense feature registration is typically used for situations where images have high similarity but lack significant structural features. Common dense registration methods include optical flow and mutual information methods, which focus on utilizing global pixel information to ensure the accuracy of the registration results.

In summary, both methods have their advantages and disadvantages. Sparse feature registration focuses more on the significance of feature points and registration efficiency, making it suitable for image registration tasks with obvious structural differences. On the other hand, dense feature registration focuses on the global details and geometric alignment of the image, and is commonly used in scenarios where high registration accuracy is required. Therefore, in practical applications, selecting the appropriate registration method requires a comprehensive consideration of the image characteristics and specific task requirements, in order to strike the best balance between accuracy and efficiency.

In addition, area-based registration methods often rely on geographic information support. By correcting distortion errors and integrating the Rational Polynomial Coefficients model of satellite images, high-precision geometric transformations and coarse geographic localization can be achieved. For UAV images, these methods need to incorporate pose information (such as GPS coordinates and attitude angles) to improve matching accuracy. However, these methods have limitations in practical applications. For instance, under complex terrain, areas with occlusions, or multi-scale image conditions, the registration accuracy and robustness may be significantly affected. Therefore, further optimization of the algorithms and the incorporation of auxiliary information are required to address the challenges in complex scenarios. Table 1 provides a comprehensive analysis of representative area-based SAR and optical image registration methods, listing the category, method improvements and conclusions of each method. The following section will provide a detailed review of the two types of area-based registration methods.

**Table 1** Summary of Area-based SAR and Optical Image Registration Algorithms

| Author | Method Category | Method Improvement | Conclusion |
|---|---|---|---|
| Ji et al., 2021 | Based on Phase Consistency and Template Matching | Phase consistency is used as a feature significance measure, and template matching algorithms are employed to match SAR and optical images. | This method effectively utilizes phase information for image matching. |
| Xiang et al., 2021 | Two-Stage Robust Registration Algorithm | In the first stage, dilation convolution features and polar rectangular template matching are used to reduce localization errors. In the second stage, bundle adjustment techniques are employed to fine-tune the correspondences. | It achieves high-precision image alignment. |
| Song et al., 2022 | New Template Matching Method: mDM-HOG | By introducing the ROEWA operator, the impact of speckle noise on gradient computation is reduced, and 3D pixel-level HOG (Histogram of Oriented Gradients) features are extracted. | It effectively enhances the robustness of template matching. |
| Zhang et al., 2023 | Optical-SAR Image Registration Framework | It includes SAR-PC-Moment for detecting sparse feature points, template matching using large local image blocks, and one-by-one registration through RANSAC. | In regions with significant terrain variations, the effectiveness and robustness of the network framework are insufficient. |
| Ye et al., 2021 | Block-Based Image Registration | Interest points are extracted, and 3D Phase Correlation (PC) is used to accelerate image matching, combined with a geometric transformation model to eliminate registration offsets. | It achieves precise registration of SAR and optical images. |
| Li et al., 2022 | New Density Descriptor Based on HOPES | It emphasizes the shape structure of image blocks, using a multi-scale Sigmoid Gabor detector and a primary edge fusion algorithm to extract edge structures. | The method is effective, but it lacks geometric invariance. |

| Yao et al., 2022 | Lightweight Feature Descriptor: MoTIF | The image gradient direction information is used to construct a diffusion tensor model, and multi-directional exponential maps are drawn through polar coordinate parameterization. | It addresses the construction of feature descriptors but does not solve the issues of scale and rotation invariance. |
|---|---|---|---|
| Ye et al., 2022 | Multi-Scale Masked Structural Feature Representation | Pixel gradient structural features are extracted at multiple scales, and a mask is constructed using large contours to reduce the influence of non-informative regions. Correspondences are obtained using an FFT-based fast template matching scheme. | It improves registration accuracy. |
| Wang et al., 2023 | Descriptor Based on Channel Direction Gradient Features | Key points are extracted by combining Block-Harris and grid points, with similarity measurement achieved through template matching. However, it cannot handle images with large rotations and scale differences. | The applicability to image matching is limited. |
| Gao et al., 2023 | Wavelet Transform and Phase Consistency-Based Frequency Domain Convolution Map Method | A multi-scale template matching strategy is employed, with most of the computations performed in the frequency domain to improve time efficiency. | It improves the matching performance for multi-modal images with displacement and scale variations, but performs poorly when both rotation and scaling are present simultaneously. |

**2.1.1 Area-based Registration Method Based on Sparse Features**

Sparse feature-based region registration methods achieve registration by extracting prominent or feature points from the image, significantly reducing computational load, and are suitable for handling images with significant structural differences. Zou et al. (2007) proposed an automated method based on the self-correlation of SAR images to determine the optimal window size for tie point matching. Ji et al. (2021) proposed a template matching method using phase consistency as a feature saliency measure, achieving matching between SAR and optical images by extracting prominent feature points. Zhou et al. (2021) used a shallow CNN model for feature extraction, focusing primarily on the significant regions of the image. These methods leverage the extraction of prominent feature points to enable efficient matching. To further reduce localization errors, Xiang et al. (2021) combined dilated convolution features with polar rectangular template matching, achieving better image alignment. Similarly, Song et al. (2022) proposed a scheme combining mDM-HOG and ROEWA operators, using three-dimensional pixel-level HOG feature matching to reduce the impact of noise and enhance the robustness of the matching process. Li et al. (2022) proposed a scheme for outlier removal using a one-class SVM, combined with grid partitioning and Harris corner detection to enhance the matching between SAR and optical images, thereby improving the robustness of the matching process. On the other hand, Zhang et al. (2023) employed SAR-PC-Moment to detect sparse feature points and utilized the RANSAC algorithm to sequentially register local regions, demonstrating superior performance in handling local region matching. Furthermore, Lei et al. (2021) proposed a GFTM (Global Feature-based Template Matching), which extracts global features from multimodal images using deep Convolutional Neural Networks, enabling rapid matching and further improving matching efficiency. Yang et al. (2024) extracted keypoints using the block-based FAST algorithm and enhanced feature descriptions by exploiting adjacent self-similarity through 3D convolution, resulting in more robust feature representations.

Sparse feature-based region matching methods are often well-suited for handling scenes with relatively clear structures and low information density due to their relatively low computational cost. These methods reduce computational complexity by extracting and utilizing prominent feature points in the image, thereby improving processing speed and efficiency. This makes them particularly advantageous in real-time applications. However, these methods may have certain limitations when dealing with tasks that require capturing fine image details. Since sparse feature methods primarily focus on salient local features, they may struggle to accurately capture small variations and complex features in images with intricate textures and rich details, resulting in reduced quality of feature extraction and ultimately affecting the matching results.

**2.1.2 Area-based Registration Method Based on Dense Features**

Dense feature-based region registration methods typically extract features at the pixel or block level in an image, capturing more detailed information. These methods are well-suited for processing images with complex textures or continuous structures.

Wu et al. (2021) extracted deep features using a Siamese U-Net to achieve matching between SAR and optical images, which is particularly suitable for multimodal matching in complex scenes. To improve matching accuracy, Ye et al. (2021) employed a block-based scheme to extract interest points, and combined 3D phase correlation with geometric transformation models for registration, resulting in higher matching precision. The histogram of oriented primary edge structure (HOPES) descriptor proposed by Li et al. (2022) uses multi-scale Sigmoid Gabor detectors and an edge fusion algorithm to extract dense features, making the matching between multimodal images more precise. Yao et al. (2022) introduced the Multi-orientation tensor index feature (MoTIF) descriptor, which constructs dense pixel-level descriptors using a diffusion tensor model, suitable for processing images with complex textures. Meanwhile, Wang et al. (2022) combined single-scale Sobel and weighted angular gradients to extract 3D dense features, and employed a non-maximum suppression strategy for matching, effectively enhancing the ability to capture image structure. Further research by Ye et al. (2022) utilized multi-scale mask structural features and employed fast Fourier transform for dense matching, which strengthened the contribution of major structural regions in the image. The CFOG-like descriptor proposed by Wang et al. (2023) combines Block-Harris and grid points for feature extraction, suitable for images with high local similarity, although it lacks robustness to large-scale rotation and scale variations. Maggiolo et al. (2022) proposed a multi-sensor optical-SAR image registration method that integrates the cGAN architecture, area-based registration framework, and COBYLA nonlinear optimization algorithm, effectively improving matching accuracy and robustness. Ye et al. (2024) used 3-D NCC (3D Normalized Cross-Correlation) for coarse matching and interest region detection, and then completed fine registration using the HOPC (Histogram of Oriented Phase Congruency ) descriptor.

Meanwhile, Cao et al. (2023) extracted deep features from multimodal remote sensing images using a pre-trained network and employed a 4D convolution-based dense template matching strategy, significantly enhancing matching efficiency. The primary structure-weighted orientation consistency (PSOC) (Li et al. , 2023) extracts primary structures using multi-scale Sigmoid Gabor filters and achieves rapid template matching through 3D Normalized Cross-Correlation. However, these methods still face challenges in matching high-resolution SAR images in complex urban environments. Gao et al. (2023) proposed a frequency-domain convolutional graph method based on wavelet transform and phase consistency, employing a multi-scale template matching strategy to improve the matching performance of multimodal images. However, the method's performance is limited in scenarios involving both rotation and scale variations. The Normalized Self-Similarity Region Descriptor (Liu et al., 2024) retains structural and directional information using Euclidean distance and cosine similarity, improving matching accuracy and consistency. Liu et al. (2024) directly performed pixel-level matching, using reference points and phase consistency modules to guide refinement in regions with significant modal differences. However, the method is sensitive to large-scale and rotational variations and requires enhanced pixel representation capabilities.

Overall, sparse feature and dense feature registration methods each have their own advantages. Sparse feature registration is computationally efficient and is suitable for processing images with prominent structures, while dense feature registration captures more details, making it ideal for handling images with complex textures and continuous structures. The choice between the two methods depends on the specific application scenario and the required registration accuracy. In the field of SAR and optical image registration, area-based registration methods continue to evolve and innovate to address challenges in different application contexts, enhancing both registration accuracy and robustness.

**2.2 Feature-Based Registration Methods**

Feature-based image registration methods are techniques that achieve image alignment by extracting prominent features from the images. These methods are widely used in fields such as medical image processing, remote sensing image analysis, and computer vision. The method primarily consists of three steps: feature detection, feature description, and feature matching.

In image registration, feature detection is first employed to extract distinctive and stable features. Then, feature descriptors are used to quantify and encode the features, capturing local texture and gradient information. Next, feature matching is performed using similarity metrics (such as Euclidean distance or Hamming distance) and matching algorithms, followed by the application of the RANSAC algorithm to remove mismatched points. A geometric transformation model (rigid, affine, or perspective) is then established to achieve precise alignment. Table 2 provides a comprehensive analysis of representative feature-based SAR and optical image registration methods, listing the category, method improvements and conclusions of each method. The following sections will provide a detailed review of the three types of feature-based methods.

**Table 2** Summary of Feature-Based SAR and Optical Image Registration Algorithms

| Author | Method Category | Method Improvement | Conclusion |
| --- | --- | --- | --- |
| Xiang et al., 2020 | Improved Phase Consistency (PC) Model | Extract uniformly distributed keypoints and select additional grid points in regions with weak structural information to extract robust feature representations. | It enhances the feature extraction capability in regions with complex structures, improving the stability of the matching |

| | | | process. |
|---|---|---|---|
| Wang et al., 2021 | Optimized SAR-SIFT Feature Registration Method | The local matching method and geometry transformation model based on the RD model are introduced, and steady-state wavelet transform (SWT) is used to optimize the SAR-SIFT keypoints. | The matching efficiency and geometric distortion handling capabilities have been improved, enhancing the system's adaptability to image structures. |
| Fan et al., 2022 | Feature extraction based on the Phase Consistency (PC) model. | Local features are extracted using the MUND-Harris detector, and a PC order-based local structural (PCOLS) descriptor is designed. | The feature points exhibit a well-distributed and highly repeatable pattern, enhancing robustness to modal variations. |
| Xiong et al., 2022 | Adjacent Self-Similarity (ASS) features. | The Adjacent Self-Similarity (ASS) features are extracted using an optimized mean offset filtering method. A feature detector and descriptor based on the minimum self-similarity map (SSM) and index map are designed. | By suppressing speckle noise, the stability and accuracy of feature extraction are enhanced. |
| Yu et al., 2021 | Algorithm based on the nonlinear SIFT framework. | By combining spatial feature detection with local frequency-domain descriptors, a multi-scale representation is constructed, and the RI-ALGH descriptor is introduced. | Robustness to local distortions has been improved, although computational efficiency still needs enhancement. |
| Li et al., 2022 | Phase Space Feature Descriptor (PSFD). | Noise is filtered and information is extracted using log-Gabor wavelet information at different frequencies. | It exhibits strong robustness to noise, but the matching efficiency still requires improvement. |
| Zhao et al., 2023 | Cher's transformation and sparse representation-based registration algorithm. | It includes the CED-HD feature point detector, multilayer complementary joint representation descriptor (MCJRD), and feature matching criterion to enhance performance. | It improves the performance of feature point extraction and description, making it suitable for registration in complex scenes. |
| Li et al., 2024 | adaptive multi-scale Phase-Only Image Feature Descriptor (AM-PIIFD) | Adaptive multi-scale extraction and feature principal orientation consistency. | Image alignment is achieved by removing unmatched KAZE features. |
| Zhang et al., 2024 | Hybrid feature-guided matching framework. | A multi-channel self-similarity map constructed during the hybrid feature coarse matching stage. | It improves feature utilization, promoting the enhancement of |
| Hou et al., 2024 | position-orientation-scale guided geometric and intensity-invariant feature transformation (POS-GIFT) | Multi-directional filtering results and position-orientation-scale guidance. | registration efficiency. The matching performance is improved through the design of robust feature descriptors. |

### 2.2.1 Feature Extraction

Feature extraction is a critical step in remote sensing image registration, directly influencing the effectiveness of subsequent feature description and registration. In recent years, researchers have proposed various improvement methods aimed at enhancing the stability, distribution, and robustness of feature points, thereby strengthening adaptability.

First, the improved Phase Consistency model (Xiang et al., 2020) enhances feature extraction robustness by uniformly extracting keypoints and supplementing grid points in regions with weak structural information. The edge and feature point detection method (Zhang et al., 2020) effectively removes inaccurate keypoints and improves detection accuracy by combining SIFT with Canny edge detection. Deep learning-based approaches (Zhang et al., 2020) utilize a twin convolutional neural network (CNN) to balance high-level semantic information with low-level detail, enhancing the hierarchical structure and deep information fusion capabilities of feature extraction. Li et al. (2017) proposed a robust feature matching method based on the Normalized Barycentric Coordinate System (NBCS).

Additionally, wavelet transform (Wang et al., 2021) optimizes SAR-SIFT keypoints, demonstrating particular advantages in handling geometric distortions. The nonlinear scale-space construction method (Yu et al., 2021) enhances feature robustness against noise and distortions through multi-scale representation, further improving feature stability by combining with the Harris-Laplace technique. RotNET (Li et al., 2021) utilizes a neural network to predict rotational relationships, enhancing the adaptability of feature points to rotation. Mohammadi et al. (2022) combined UR-SURF (Uniform Robust-Speeded Up Robust Features) feature extraction, RISS (Rotation Invariant Self-Similarity) descriptors, LWGTM (Localized Weighted Graph Transformation Matching) outlier rejection, and the TPS (Thin-Plate Spline) transformation model for image registration.

In local feature extraction, the OGGS operator (Li et al., 2021) improves the accuracy of rotational differences by calculating the absolute directional derivative histogram. Fan et al. (2022) combine phase consistency with the Harris detector to ensure uniform distribution and high repeatability of feature points across the entire image. Methods based on self-similarity features (Xiong et al., 2022) and multi-scale fusion strategies (Liu et al., 2023) further enrich the dimensionality of feature extraction in remote sensing images. Ren et al. (2024) employ a side-window filter (SWF) to construct a new scale space, improving the Harris algorithm for keypoint extraction and using logarithmic polar coordinate descriptors to enhance matching quality. Yang et al. (2023) achieved robust and efficient image matching by combining a feature extractor based on both spatial domain scale space and frequency domain scale space.

The primary innovation direction in feature extraction techniques focuses on enhancing the repeatability and robustness of feature points, ensuring consistency across different conditions and providing stronger adaptability. To achieve this, multi-scale analysis, effective utilization of edge information, and integration with deep learning technologies have significantly strengthened the robustness of feature extraction methods. For example, multi-scale analysis captures detailed variations of the image at different scales, allowing for better handling of scale changes and partial occlusions. Edge information helps identify prominent structures, enhancing the discriminative power of features. The introduction of deep learning provides stronger non-linear expression capabilities, making feature descriptions richer and more effective.

Despite these innovations achieving significant success in many applications, the accuracy of feature extraction remains a challenge for images with limited structural information. The absence of distinct edges, textures, or geometric features makes it difficult for traditional feature point detection methods to identify stable keypoints, thereby affecting subsequent image matching and recognition performance. Even with the use of deep learning models, the scarcity of structural information may lead to a lack of prominent patterns during the feature learning phase, resulting in insufficient feature discriminability. Therefore, effectively extracting meaningful feature points from such images remains a critical issue in the current development of feature extraction techniques. This calls for more innovative strategies, such as integrating contextual information or employing feature enhancement techniques, to improve the overall performance of feature extraction.

### 2.2.2 Feature Description

Feature description refers to the process of quantifying and encoding the extracted feature points, enabling them to be matched in subsequent image processing and computer vision tasks. This process is critical within the entire visual system, as its performance directly affects the accuracy and efficiency of feature matching. The core challenge in feature description is to effectively represent the key features in an image, ensuring that these features maintain strong discriminability and consistency under variations in viewpoint, scale, lighting, and noise interference. Therefore, the performance of feature description directly determines the success rate of image matching, as well as the overall speed and accuracy of the system's processing.

Based on the improved Phase Consistency model, Xiang et al. (2020) extract robust feature representations, enhancing adaptability to modal variations. Zhang et al. (2020) optimized the SIFT descriptor using the FLANNs algorithm to improve matching performance. Zhang et al. (2020) employed convolutional neural networks to extract deep dense features, enhancing the descriptive capability for complex scenes.

To address rotation and scale invariance, Yu et al. (2021) proposed the rotation-invariant amplitudes of log-Gabor orientation histograms (RI-ALGH) descriptor, which enhances robustness under local distortions. The Phase Space Feature Descriptor (Li et al., 2022) processes noise information using log-Gabor wavelets at different frequencies, improving the descriptor's performance in noisy environments. The PCOLS descriptor proposed by Fan et al. (2022) encodes local structures through grouping, enhancing the feature's expressive capability.

Additionally, the MCJRD descriptor by Zhao et al. (2023) combines Cher's transformation and sparse representation, enhancing the descriptive performance of feature points. Cui et al. (2020) achieved rotation invariance in feature point descriptions through a rotation-invariant coordinate system. Yu et al. (2021) designed a novel consistent feature map and structural descriptor, further improving matching accuracy. Lv et al. (2024) identified significant structures using Window Inherent Variations (WIVs) and employed a Maximum Response Index (MRI) filter bank to extract multi-directional structural features, constructing a radial invariance descriptor.

In summary, recent innovations in feature description have primarily focused on continuously improving the robustness of descriptors, particularly in addressing complex scenarios such as rotation, scale variations, and multi-modal changes. Ensuring that descriptors maintain high reliability and consistency under these conditions has become a key area of research. These innovations enable descriptors to accurately describe feature points and achieve stable registration, even in the face of challenges such as viewpoint changes, varying lighting conditions, and noise interference. Future research can further explore the design of lightweight and efficient feature descriptors to meet the stringent computational resource constraints and speed requirements in practical applications.

**2.2.3 Matching Strategies**

Matching strategies are built upon feature descriptions, where feature points are matched to achieve image registration or alignment. The goal is to identify corresponding features across different images, thereby combining the two images to form a more comprehensive view or enable subsequent visual tasks. In recent years, research on matching strategies has focused on improving matching efficiency, accuracy, and adaptability to complex transformations. Zhang et al. (2020) utilized the FLANNs algorithm to achieve efficient matching, significantly reducing matching time. Wang et al. (2021) proposed a strategy that combines local matching with a geometric transformation model, enhancing the ability to handle geometric distortions. The weighted strategy based on optical flow energy functions (Yu et al., 2022) integrates SIFT and PC descriptors, mitigating the impact of noise and improving matching accuracy.

The FFT-accelerated SSD matching method (MaskMIND, Yu et al., 2022) accounts for position uncertainty, significantly improving matching speed. The sparse representation matching criterion (Zhao et al., 2023) optimizes feature matching through local constraints, greatly enhancing matching performance. Li et al. (2024) eliminated mismatched points by ensuring consistency in feature principal orientations, ensuring the reliability and accuracy of the matching process. Innovations in matching strategies also include the hybrid feature-guided matching framework (Zhang et al., 2024), which improves matching efficiency through a coarse-to-fine optimization approach. POS-GIFT (Hou et al., 2024) enhances inlier recovery capability with a multi-dimensional guiding strategy, significantly boosting matching performance. Xiong et al. (2024) introduced Gaussian smoothing, neighborhood downsampling, and Position Weighted Matching (PWM) strategies based on the classic OSS (Oriented Self-Similarity) descriptor to improve robustness and matching efficiency. GLS-MIFT (Fan et al., 2024) generates feature maps through multi-scale, multi-directional Gaussian filtering, extracting descriptors invariant to rotation, scaling, and translation, and uses a global-to-local search strategy to enhance matching reliability.

In summary, innovations in matching strategies have primarily focused on improving matching efficiency and accuracy. As the scale of image data continues to expand, an important direction for future research will be to further reduce matching time while maintaining high precision. Additionally, developing matching strategies with high adaptability for different application scenarios is a critical future challenge. For instance, in highly dynamic scenes with complex geometric transformations, designing matching strategies that are more robust to lighting, noise, and occlusions can significantly improve matching success rates. Ensuring high matching accuracy while achieving fast and precise feature matching in complex environments will become a key challenge and innovation goal in the development of visual algorithms.

Feature-based registration methods have made significant progress in feature extraction, feature description, and matching strategies. Innovations in feature extraction mainly focus on improving the stability and distribution of feature points, while feature description emphasizes enhancing the robustness and discriminability of descriptors. Matching strategies, on the other hand, concentrate on improving matching efficiency and accuracy. However, feature-based matching methods still face challenges in handling low-structure images, addressing rotation and scale variations, and performing well in complex backgrounds. These issues are particularly prominent in the registration of SAR and optical images. Future research could further explore the integration of multi-modal information and deep learning technologies to improve the generalization ability of SAR and optical image registration. For instance, by utilizing deep learning models for feature fusion from multiple sources, the discriminability and matching performance of features could be effectively enhanced. In terms of real-time processing and efficiency, the development of lightweight feature extraction and description algorithms will be crucial for achieving efficient registration in resource-constrained environments. Additionally, for SAR images, which often lack texture features, exploring ways to better extract and utilize their unique structural and backscatter information remains a key area for future research. These directions will provide more robust and precise solutions for the fusion applications of SAR and optical images.

### 2.3 Deep Learning-Based Registration Methods

Deep learning-based registration methods for SAR and optical remote sensing images are one of the key research directions in the field of remote sensing image processing. These methods aim to effectively address the registration challenges arising from the significant differences in imaging principles, visual representations, and geometric distortions between SAR and optical remote sensing images. Currently, approaches to solving the SAR and optical image registration problem can be classified into four categories: modality-unified methods, feature extraction-based registration methods, matching strategy-based methods, and end-to-end deep learning-based solutions.

Firstly, modality-unified methods aim to reduce the differences between SAR and optical images by converting them into a common modality, thereby achieving a unified feature representation. These methods often employ technologies such as Generative Adversarial Networks (GANs) to transform SAR images into optical-style images, or vice versa, to facilitate more direct feature matching. This modality transformation helps bridge the gap between the two image types to some extent, but during the style transfer process, some original information may be lost, potentially affecting the registration accuracy.

Secondly, feature extraction-based matching methods leverage the powerful feature learning capabilities of deep learning networks to separately extract prominent features from both SAR and optical images, enabling cross-modal feature alignment. The key to this approach is to identify common features that are suitable for both image modalities, thereby enabling an effective mapping of the feature space. While this method has made progress in feature description, challenges remain in terms of matching accuracy and stability, especially in complex scenes where there are significant differences in image texture and geometric structure.

Next, matching strategy-based methods achieve effective registration of SAR and optical images by designing and optimizing matching algorithms. Innovations in matching strategies are crucial for improving feature utilization efficiency and registration accuracy. Researchers have made continuous improvements in matching metrics, outlier rejection, and overall algorithm design, enabling high-precision and widely applicable registration results that can handle complex scenarios. However, this approach still faces several challenges in practical research and applications, including difficulties in cross-modal feature extraction and matching, the complexity of outlier rejection, high computational complexity, and reliance on deep learning models. To achieve more efficient and robust registration, research must strike a reasonable balance between feature matching accuracy and computational efficiency, while further innovating and optimizing matching strategies.

Finally, end-to-end deep learning frameworks have gradually become a research hotspot. By constructing an end-to-end network architecture, the SAR and optical image registration problem can be directly modeled as a learning task, enabling a fully automated registration process from input to output. The advantage of such methods is that they can leverage large amounts of labeled data for supervised learning, effectively capturing cross-modal feature relationships in complex scenes. However, the performance of these models relies heavily on vast amounts of high-quality training data, and obtaining sufficient labeled data in practical applications remains a significant challenge.

Despite the distinctive characteristics of these methods and their varying degrees of success in improving registration accuracy and robustness, challenges remain when addressing complex scenarios, such as dramatic terrain changes, significant seasonal differences, and variations in lighting and imaging conditions. These challenges demand further solutions. Therefore, future research may focus on developing algorithms with enhanced robustness and generalization capabilities to address cross-modal remote sensing image registration in diverse and complex environments. Moreover, there is a need to further improve the automation and precision of the registration process. The application of deep learning in remote sensing image matching has significantly boosted registration accuracy and robustness. Ongoing research continues to innovate in areas such as feature extraction, matching strategies, end-to-end methods, and modality unification, providing rich theoretical and methodological support for addressing remote sensing image registration challenges in complex scenarios. Table 3 offers a comprehensive analysis of representative methods for deep learning-based SAR and optical image registration, listing the category, method improvements and conclusions of each method. The following section will provide a detailed review of the four aforementioned deep learning-based registration approaches.

**Table 3** The Summary of Deep Learning-Based SAR and Optical Image Registration Algorithms

| Author | Method Category | Method Improvement | Conclusion |
|---|---|---|---|
| Hughes et al., 2018 | Generative Adversarial Networks (GAN) and Variational Autoencoders (VAE) | Using GANs and Variational Autoencoders (VAE) to generate novel SAR image patches for constructing a balanced training dataset. | The quality of generated samples still needs to be improved. |
| Du et al., 2021 | CycleGAN | The application of unsupervised image synthesis in SAR-optical image matching was studied, with the performance enhanced through shared matching strategies.。 | There is a need to develop better mapping strategies, such as designing generator architectures based on the imaging mechanisms. |
| Zhang et al., 2023 | Feature Adversarial Network | Using generative adversarial learning to map images of different modalities to a shared subspace, and optimizing the features through a modality discriminator to make them indistinguishable across modalities. | The matching results are not sufficiently stable in certain cases and are dependent on the quality of the generated images. |
| Lin et al., 2023 | CNN-Transformer | By using dense blocks and transition | It improves feature |

| | hybrid feature descriptor | layers to maximize gradient differences, this approach avoids information redundancy and achieves a balance between performance and computational burden. | extraction performance while controlling computational overhead. |
|---|---|---|---|
| Xiang et al., 2022 | Feature Decoupling Network (FDNet) | It consists of RDNet and pseudo-Siamese fully convolutional network (PSFCN), which are used to learn noise features and semantic features, respectively, enhancing the robustness of the feature extraction process against noise. | It enhances the robustness of feature extraction against noise. |
| Xu et al., 2023 | a local descriptor for SAR-optical matching (SODescNet descriptor) | The SOPatch dataset was created, and the SODescNet descriptor was proposed. Through pre-training with a Siamese network and fine-tuning with a pseudo-Siamese network, the feature extraction capability was validated across different modalities and complex scenarios. | It improves the feature extraction capability in different modalities and complex scenarios. |
| Zhang et al., 2022 | Siamese Domain Adaptation Method | Based on a combined loss function, this method improves unsupervised matching performance through self-learning and multi-resolution histogram matching, making it particularly suitable for situations where labeled data is scarce. | It enhances the adaptability in scenarios with a lack of labeled data. |
| Chen et al., 2023 | Co-Attention Matching Module (CAMM) | It integrates the structural features of image pairs while optimizing the processing efficiency of large images, effectively addressing the issue of high computational cost. | It improves the efficiency of large image processing and reduces computational costs. |
| Li et al., 2023 | Semantic Position Probability Distribution Matching Method | By combining the ResNet-50 + FPN network to generate multi-scale features, it demonstrates high accuracy and robustness in cross-modal image matching. | It enhances the accuracy and robustness of cross-modal image matching. |
| Li et al., 2023 | Dual-Branch Cross-Fusion Network (DF-Net) | By integrating deep learning techniques, end-to-end matching of cross-modal images is achieved, significantly improving matching accuracy and efficiency. | It improves matching accuracy and efficiency. |
| Deng et al., 2022 | Recoupling detection and | Through mutual weighting strategies and super detectors, end-to-end feature | It enhances the registration performance |

| | description(ReDFeat) | detection and description are provided, enhancing the overall performance of multi-modal image registration. | of multi-modal images. |
| --- | --- | --- | --- |
| Chen et al., 2022 | Hierarchical Consistency Attention Network (HCA-Net) | By adaptively enhancing neighborhood consistency, end-to-end matching is achieved, making it suitable for handling high-complexity image scenes. | It enhances the matching quality in high-complexity scenes. |

**2.3.1 Methods Based on Modality Difference Elimination**

Modal unification methods have gradually become an effective approach to solving the SAR and optical image registration problem in recent years. The core idea is to reduce the differences between SAR and optical images by converting them into the same style or modality, thereby achieving a unified feature representation. These methods typically utilize technologies such as Generative Adversarial Networks (GANs) to convert SAR images into optical-style images, or vice versa, enabling more direct feature matching.

Hughes et al. (2018) proposed using Generative Adversarial Networks (GANs) and Variational Autoencoders to generate novel SAR image patches for constructing a balanced training dataset. This generative framework aims to address the issue of imbalanced training data, but the quality of the generated samples still requires improvement. Maggiolo et al. (2020) introduced a method combining cGAN, area-based l2 similarity metric, and the COBYLA algorithm to achieve multi-sensor image registration, demonstrating that the correlation metric combined with cGAN is effective in multi-sensor registration. However, the method depends on the quality of GAN-generated images, requiring further optimization. Du et al. (2020) proposed the K-means Clustering Guide Generative Adversarial Networks (KCG-GAN), combining feature matching loss, segmentation loss, and L1 loss to improve the accuracy of SAR to optical image matching. Future work can explore more complex segmentation methods to improve generalization. Du et al. (2021) studied the application of unsupervised image synthesis in SAR-optical image matching, using CycleGAN to enforce feature matching consistency and improving matching results through shared matching strategies. However, better mapping strategies need to be developed, such as designing generator architectures based on imaging mechanisms. Nie et al. (2022) introduced a dual-generator transformation network, combining a texture generator and a structure generator, merging structural and texture features using frequency-domain and spatial-domain loss functions to achieve high-precision transformation results. However, the model's generalization ability still needs further enhancement. Zhang et al. (2023) proposed the Feature Adversarial Network for multi-modal template matching, using generative adversarial learning to map images from different modalities into a shared subspace and optimizing these features through a modality discriminator to make them indistinguishable, thus enabling effective correlation learning. However, this method relies on the quality of the generated images, resulting in unstable matching results in certain cases. James et al. (2023) presented an image registration method based on Conditional Generative Adversarial Networks (cGAN) and Fast Fourier Transform (FFT) correlation, using class-optical patches generated from SAR image patches to achieve precise registration between SAR and optical images. RSENet (Nie et al., 2024) includes the HOG-GAN image translation module and an equivariant descriptor module, achieving rotation and scale invariance through Scale Angle Estimation (SAE) and Periodic Encoding (PE).

Although modality unification methods based on style transfer have made some progress in cross-modal image registration, they still have several shortcomings. For example, the quality of the generated images can be unstable, and style transfer may lead to information loss, which ultimately affects the registration accuracy. Therefore, future research should focus on further improving the quality of GAN-generated images to minimize information loss and enhance registration accuracy and robustness. Additionally, developing more robust style transfer algorithms to address complex imaging conditions and the diverse demands of different scenarios is an important topic for future exploration.

**2.3.2 Feature Extraction**

Feature extraction-based matching methods are one of the key approaches to solving the SAR and optical image registration problem. The core idea is to leverage the powerful feature learning capabilities of deep learning networks to separately extract prominent features from both SAR and optical images, achieving cross-modal feature alignment. Many researchers have conducted in-depth studies on feature extraction, aiming to obtain feature representations that are both highly discriminative and have strong generalization capabilities.

Liu et al. (2023) proposed the Global-Local Consistency Network (GLoCNet), which combines global transformation consistency with local neighborhood consistency modules. By dynamically adjusting the upper limit distance for outliers, it enhances the stability and robustness of feature extraction. Chen et al. (2023) designed the Shape-Former network, which integrates CNN and Transformer architectures, using the permutation-invariant ShapeConv operation to capture local shape details of sparse data, thereby improving the flexibility and accuracy of feature extraction. Lin et al. (2023) introduced the CNN-Transformer hybrid feature descriptor, maximizing gradient differences through dense blocks and transition layers to avoid information redundancy, achieving a balance between performance and computational burden. Hu et al. (2023) proposed the Modality Shared Attention Network (MSA-Net), which utilizes modality-specific and modality-shared feature learning modules to enhance the quality of cross-modal feature representations between SAR and optical images. Li et al. (2022) developed the Cross-Modal Feature Description Matching Network (CM-Net), which improves cross-modal image matching performance through context-aware self-attention and cross-fusion modules. Chen et al. (2022) introduced CSR-Net, which combines a permutation-invariant structural representation learning module with a context-aware attention mechanism, effectively enhancing fine-grained pattern recognition and making the feature extraction process more stable and robust.

Additionally, Liao et al. (2021) proposed MatchosNet, which utilizes local deep feature descriptors and a hard L2 loss function to optimize the precision of feature extraction. Chen et al. (2021) introduced the Hierarchical Local Structure Visualization (HLSV) method, which decouples the local structures of feature points through hierarchical tensor decomposition and incorporates permutation-invariant layers (PIL) to enhance the model's generalization ability. Xiang et al. (2022) proposed the Feature Decoupling Network (FDNet), composed of RDNet and PSFCN, which are used to learn noise features and semantic features, respectively, further enhancing the robustness of the feature extraction process against noise. These methods have made groundbreaking contributions to the details and generalization ability of feature extraction, improving feature representation across different scenarios. Xu et al. (2023) created the SOPatch dataset and proposed the SODescNet descriptor, which is pre-trained using a Siamese network and fine-tuned using a pseudo-Siamese network, validating the feature extraction capability in different modalities and complex scenarios. Li et al. (2022) proposed the Patch Matching Network (PM-Net), which extracts robust keypoints through a multi-level keypoint detector (MKD), improving the reliability of feature extraction. Xiang et al. (2023) introduced a feature intersection-based keypoint detector (FI), combining phase consistency, gradient orientation, and local variation coefficient detectors to extract highly repeatable keypoints, further enhancing adaptability to high-complexity images. Wu et al. (2024) combined multiple feature extraction strategies with Graph Neural Networks (GNN), generating grid feature points using a no-detector strategy and improving registration accuracy through a clustering-based two-stage framework. Ye et al. (2024) used a multi-branch global attention (MBGA) module to enhance feature representation and optimized positive-negative sample differentiation ability through multi-crop image matching loss (MCTM). GLFE-Net (Zhao et al., 2024) is a dual-branch network that combines a Transformer to extract global features and a CNN to extract local features, optimizing the discriminative ability of descriptors through a hard mean loss.

Although these methods have made some progress in feature description, the accuracy and stability of registration still face challenges in complex scenarios due to significant differences in image texture and geometric structure. Future research should focus on developing more robust and discriminative feature extraction mechanisms to improve registration accuracy and generalization, especially in complex and dynamic environments.

### 2.3.3 Matching Strategies

Innovations in matching strategies are crucial for improving feature utilization efficiency and registration accuracy. In recent years, researchers have continuously improved matching metrics, outlier removal techniques, and overall algorithm designs to enhance image matching efficiency and precision. Below is a summary of some innovative methods in the matching domain that have achieved significant results and their impact on practical applications.

First, in terms of matching metrics, Hoffmann et al. (2019) proposed a registration method based on fully convolutional networks, which significantly improves registration accuracy compared to traditional mutual information-based registration metrics. Hu et al. (2023) introduced I2MGAN, which combines a contrastive learning framework and explores feature similarity, greatly improving multi-modal image matching, especially in complex image scenarios. RSOMNet (Zhang et al., 2024) employs a multi-granularity matching strategy and a Non-Shared Content Filtering (NSCF) module, first narrowing the search range using coarse-grained features and then precisely locating matching positions with fine-grained features.

In terms of outlier rejection and optimization, Hughes et al. (2020) developed a three-step deep learning framework that generates correspondence heatmaps through cross-correlation in multi-scale feature spaces and optimizes the matching process by removing outliers, significantly improving matching performance in complex terrains. Chen et al. (2021) introduced adaptive dual-aggregation convolution and point rendering layers in StateNet to enhance outlier suppression, improving matching accuracy in complex transformation scenarios. Additionally, Chen et al. (2022) proposed the Hierarchical Consistency Attention Network (HCA-Net), which adaptively enhances neighborhood consistency, effectively removing mismatched points and significantly improving overall matching quality. Li et al. (2022) combined nearest-neighbor search with the MAGSAC++ algorithm, greatly enhancing outlier rejection efficiency, particularly in scenarios with a large number of outliers. Liaghat et al. (2023) used deep neural networks to distinguish between correct and incorrect matches, effectively reducing erroneous correspondences and significantly enhancing the robustness of overall matching.

In terms of overall algorithm design, Zhang et al. (2022) proposed a Siamese domain adaptation method based on a combined loss function, which enhances unsupervised matching performance through self-learning and multi-resolution histogram matching, demonstrating exceptional adaptability, especially in the absence of labeled data. Zhang et al. (2023) further introduced an extended feature concatenation method and a self-supervised optical flow fine-tuning strategy, combining dense and sparse registration to effectively improve matching accuracy, particularly for handling large-scale remote sensing data. Chen et al. (2023) proposed the Co-Attention Matching Module (CAMM), which integrates the structural features of image pairs while optimizing the processing efficiency of large images, effectively addressing the issue of high computational cost. Dou et al. (2023) used the DDL network in the fine-registration stage to handle image modality variations, further enhancing multi-modal image registration performance. Additionally, Li et al. (2023) introduced a matching method based on semantic position probability distributions, which generates multi-scale features by combining the ResNet-50 + FPN network, showing high accuracy and robustness in cross-modal image matching.

In summary, innovations in matching strategies have made significant progress in improving image registration accuracy, reducing the impact of outliers, and optimizing overall algorithms. These innovations, through the introduction of deep learning and adaptive mechanisms, have greatly enhanced matching performance in complex scenarios, particularly in multi-modal and cross-modal environments. However, current methods still have some shortcomings, such as high computational resource demands and limited generalization in specific application scenarios. Future research can focus on reducing computational complexity and improving the robustness and versatility of algorithms to better meet the demands of practical applications.

**2.3.4 End-to-End Registration Methods**

With the rapid development of deep learning technology, end-to-end methods have gradually become an important tool in the field of image registration. These methods integrate feature extraction and the registration process, providing a holistic solution that makes image registration more automated and efficient, attracting increasing attention from researchers.

Zeng et al. (2020) proposed a CNN-based automatic remote sensing image region registration method, which achieves pixel-level one-to-one correspondence output, significantly simplifying the registration process and improving registration accuracy. Xu et al. (2021) introduced Him-Net, which optimizes matching results through a heatmap loss function and realizes end-to-end template matching, significantly reducing human intervention in complex image matching. Fang et al. (2021) proposed an end-to-end deep learning model for SAR and optical images based on Siamese U-net and Fast Fourier Transform, greatly enhancing the automation of registration and effectively addressing issues of rotation and scale variation. Du et al. (2022) combined Pix2pix and CycleGAN to propose a semi-supervised image-to-image translation framework, providing a flexible end-to-end solution that performs exceptionally well in handling image modality changes. Siamese domain adaptation method (Zhang et al., 2022) and Dual-Branch Cross-Fusion Network (DF-Net) (Li et al., 2023) both integrated deep learning technologies to achieve end-to-end matching of cross-modal images, significantly improving matching accuracy and efficiency. Sun et al. (2024) combined symmetry loss to generate confidence masks, select reliable optical flow points, and estimate affine transformation parameters through linear regression, but their method has limited adaptability to large occlusions and large-scale, rotational transformations.

Additionally, Liao et al. (2021) proposed the MatchosNet algorithm, which utilizes local deep feature descriptors for efficient matching, achieving an end-to-end matching process that significantly improves matching efficiency and accuracy. Li et al. (2022) introduced CM-Net, which transforms the matching task into a classification task, optimizing feature representation and thus enabling end-to-end matching while reducing the complexity of multiple steps. Chen et al. (2021) developed CSR-Net, which enhances fine-grained pattern recognition through an end-to-end approach, making the registration process more accurate and efficient. Cui et al. (2021) proposed MAP-Net, which combines convolutional neural networks, attention mechanisms, and spatial pyramid pooling to achieve end-to-end image matching, making it particularly suitable for multi-scale remote sensing image processing. Deng et al. (2022) introduced the ReDFeat method, which provides end-to-end feature detection and description through mutual weighting strategies and super detectors, significantly enhancing the overall performance of multi-modal image registration. Chen et al. (2022) proposed the Hierarchical Consistency Attention Network (HCA-Net), which adaptively enhances neighborhood consistency, achieving an end-to-end matching process that is particularly effective for handling high-complexity image scenes.

In summary, end-to-end deep learning methods have demonstrated great potential in the field of image registration, seamlessly integrating feature extraction and the matching process, significantly improving registration efficiency and accuracy. Their advantages include high automation, enhanced precision, and adaptability to complex scenarios. However, current end-to-end methods still require improvement in dealing with registration under extreme conditions, particularly in multi-modal, high-noise, and extreme scale variation situations. Additionally, the interpretability of the model and its computational resource requirements are important factors limiting further development. Future directions may include enhancing the model's robustness to complex environments, incorporating more prior knowledge to improve model interpretability, and reducing hardware resource demands through lightweight designs. Furthermore, the further exploration of multi-modal fusion technology is expected to play a greater role in end-to-end image registration.

Significant progress has been made in the application of deep learning to remote sensing image matching, with continuous innovations in methods for feature extraction, matching strategies, end-to-end solutions, and style transfer techniques, significantly improving registration accuracy and robustness. However, current methods still face challenges when dealing with complex scenarios, such as high computational complexity and sensitivity to strong radiometric differences between cross-modal images. Future research should focus on designing more efficient network architectures to reduce computational costs and enhance the model's generalization capabilities. At the same time, integrating multi-modal, multi-scale, and multi-stage fusion methods is expected to further improve the robustness and applicability of remote sensing image registration. In conclusion, deep learning-based matching methods offer new insights into remote sensing image registration, and future research could continue to drive the development of this field by incorporating advanced technologies such as self-supervised learning, Generative Adversarial Networks (GANs), and multi-modal information fusion.

## 3. Developments in the Registration of High-Resolution SAR and Optical Remote Sensing Images

The development of high-resolution SAR image registration has introduced higher demands for detail and accuracy in multi-modal remote sensing image processing, while also presenting new technological challenges. At high resolution, the fine targets, texture features, and geometric details in the image become clearer and more pronounced, which requires higher precision for image registration. However, due to differences in imaging mechanisms between different sensors, significant non-linear radiometric differences exist between SAR and optical images. The multiplicative speckle noise in SAR images, deeply entangled with robust structural features, often hides the useful information, making the identification of dense homologous points particularly challenging, which further complicates image registration. In addition, the complex terrain undulations and geometric distortions in high-resolution SAR images pose significant challenges for precise registration. Therefore, how to effectively handle these complex noises and geometric distortions while extracting stable structural features has become a key issue in high-resolution SAR image registration. In recent years, high-resolution SAR image registration technology has made innovations in several areas to improve registration accuracy and robustness. Table 4 provides a comprehensive analysis of representative methods for high-resolution SAR-optical image registration, listing the category, method improvements and conclusions of each method.

**Table 4** Summary of High-Resolution SAR-Optical Remote Sensing Image Registration Methods

| Author | Method Category | Method Improvement | Conclusion |
| --- | --- | --- | --- |
| Xiang et al., 2018 | OS-SIFT Algorithm | The OS-SIFT algorithm uses Harris scale-space and multi-scale Sobel operators for keypoint detection, orientation assignment, and descriptor extraction. | It achieves robust registration, but performs poorly in urban areas. The performance can be improved by integrating regional methods, such as mutual information. |
| Merkle et al., 2018 | Conditioned GAN-Based Matching | Generating SAR-like image patches from optical images to improve matching accuracy. | The reliability of the generator network in preserving the geometric structure of optical images needs further improvement. |
| Zhang et al., 2021 | OSMNet (Multi-level Feature Fusion) | Multi-frequency channel attention, Adaptive Weight Loss (SAW), high localization accuracy. | Performs excellently in optical and SAR image matching; still requires improvement in feature point detection and outlier removal. |
| Hughes et al., 2020 | End-to-End Sparse Matching Framework | The Goodness network is used for proposal generation, cross-correlation heatmaps, and outlier reduction. | Compared to NCC and pseudo-Siamese methods, the accuracy is significantly improved; however, it is limited by the Goodness network architecture and inter-domain offset handling. |
| Paul et al., 2021 | Mutual Information + SPSA Optimization | Using mutual information and SPSA for coarse-to-fine registration to estimate global translation differences. | Effective for globally corrected images; however, it faces challenges when handling images with significant global geometric differences. |
| Ye et al., 2022 | Multi-scale Masked Structural Feature Matching | Extracting pixel gradient features at multiple scales and using FFT-based template matching. | Robustness is improved when handling uniform regions; however, integrating more structural features (such as HOPC, PCSD) is needed to enhance performance. |
| Zhang et al., 2022 | Siamese Domain Adaptation | Combined loss function, using rotation/scale-invariant transformation modules, and multi-resolution histogram matching. | Significantly improves domain adaptation and matching performance; requires enhancement in handling non-rigid registration parameters. |

| Xiang et al., 2024 | Detector-free Feature Matching | Two-step strategy, downsampling descriptor matching, and pseudo-epipolar constraints. | Achieves high computational efficiency through GPU parallelization; adaptability in complex scenes needs improvement. |
|---|---|---|---|
| Wang et al., 2024 | Gray-Level Normalized Mutual Information | Block-based registration strategy to balance issues of insufficient information and excessive overlap. | Achieves fast automatic registration, however, it has limitations when handling high local variations. |
| Xiang et al., 2023 | Global-to-Local Registration Algorithm | Geographic coding, global matching, local matching, and multidirectional anisotropic Gaussian derivatives (MAGD) features. | Performs well when handling large shifts; needs to incorporate thin-plate spline models to address non-linear misalignment issues. |

### 3.1 Feature Enhancement Matching

Merkle et al. (2018) explored a multi-sensor image matching method based on Conditional Generative Adversarial Networks (cGAN), generating SAR-like image patches from optical images to improve the matching accuracy between SAR and optical images. By generating simulated images, this method enhances feature alignment and matching performance between multi-modal images. Additionally, Ye et al. (2022) proposed a multi-scale mask-based structural feature matching method, which extracts structural features at multiple scales and uses masks to enhance the contribution of the primary structural regions, thereby significantly improving the robustness of image matching.

### 3.2 Optimization of Non-Rigid Deformation Handling

There are often significant nonlinear geometric and radiometric differences between SAR and optical images, making non-rigid deformations a challenging issue in image registration. Zhang et al. (2022) proposed a Siamese domain adaptation method combined with a causal inference model, addressing the nonlinear deformation issues in images through multi-resolution histogram matching and rotation/scale invariant modules. This method, through a self-learning strategy, enables the model to better align across different domains.

### 3.3 Image-based Generation of Matching Optimization

Merkle et al. (2018) used Conditional Generative Adversarial Networks (cGAN) to generate SAR-like image patches, providing a new approach for improving the alignment of images from different modalities. This method not only effectively captures features during the generation of simulated images but also enhances matching accuracy through the alignment of the generated images. Ye et al. (2021) also achieved joint alignment of multi-modal images through multi-scale feature extraction and mask enhancement.

### 3.4 Registration of Large-Scale Scenes

Zhang et al. (2022) demonstrated the superior registration performance of their Siamese domain adaptation method through experiments on the RadarSat/Planet and Sentinel-1/2 datasets. The method effectively handles image registration tasks across large-scale scenes, even with varying resolutions and imaging conditions. Additionally, the end-to-end matching framework proposed by Hughes et al. (2020) has proven its applicability in large-scale scenes. By generating candidate patches with the Goodness network, it significantly improves the matching efficiency between SAR and optical images across different scenarios. Multi-Scale Description of Detailed Features: High-resolution images contain rich geometric and texture details, and traditional single-scale feature descriptors are often inadequate for high-resolution data. Current research is shifting toward multi-scale feature extraction methods to capture image details across multiple scales. By integrating features from different scales, registration algorithms achieve a better balance between detail accuracy and global stability. Zhang et al. (2021) introduced a new high-resolution optical and SAR image matching network (OSMNet), which utilizes multi-level feature fusion and multi-frequency channel attention modules to enable effective cooperation of different types and levels of image features, balancing detail accuracy with global robustness. However, challenges remain in feature point detection and handling outlier control points, which require further improvement to enhance registration accuracy.

### 3.5 Deep Learning-Based Feature Representation

Deep neural networks have demonstrated strong adaptability in multi-modal feature alignment and learning. The introduction of self-supervised and unsupervised learning methods, in particular, enables models to automatically learn aligned feature representations on large-scale datasets, reducing the reliance on a large amount of manually labeled data. Zhang et al. (2022) proposed an unsupervised Siamese Domain Adaptation (SDA) method, which reduces the domain gap between SAR and optical datasets through a combination loss function. The method uses rotation/scale-invariant transformation modules to extract features, significantly improving registration performance. However, there are still shortcomings in handling non-rigid registration parameters, which could be improved by considering non-rigid deformations between SAR and optical images to further enhance matching accuracy. Additionally, the end-to-end matching framework proposed by Hughes et al. (2020) incorporates self-supervised learning and uses the Goodness network to generate candidate patches with high matching probabilities, leading to improvements in both accuracy and precision. However, the outlier reduction mechanism in the matching process still requires further improvement.

### 3.6 Optimization of Geometric Distortion and Noise Handling

High-resolution SAR images are often affected by significant geometric distortions and noise due to complex terrain variations, viewing angles, and electromagnetic wave scattering characteristics. To improve registration accuracy, existing methods have incorporated precise terrain modeling and geo-referenced matching algorithms for multi-modal geometric distortion correction, enhancing the spatial consistency between high-resolution SAR and optical images. Xiang et al. (2023) proposed a global-to-local registration algorithm that uses multidirectional anisotropic Gaussian derivatives (MAGD) features to enhance noise resistance and geometric consistency. By combining geographic coding with global matching, this method reduces the impact of terrain undulations on the matching process. However, the method still requires further adjustment when dealing with non-linearly misaligned images, and integrating thin-plate spline models or piecewise linear models could help address complex deformation issues. Paul et al. (2021) introduced a registration method that uses mutual information and SPSA optimization to achieve a coarse-to-fine matching strategy. However, the method still faces difficulties when matching images with significant geometric differences. The matching algorithm could be improved to better handle substantial geometric discrepancies.

### 3.7 Development of Real-Time Registration Methods

With the increasing frequency and resolution of remote sensing data acquisition, the demand for real-time registration of high-resolution SAR images has become increasingly important. To achieve an efficient registration process, high-performance computing acceleration methods and lightweight network architectures have been introduced, resulting in higher registration speeds and providing strong technical support for practical applications. Xiang et al. (2024) proposed a detector-free feature matching method that adopts a two-step strategy, achieving near-real-time computational efficiency through GPU parallel processing, significantly improving registration efficiency and accuracy. However, its adaptability in complex scenarios still needs enhancement, which can be addressed by further optimizing the algorithm to handle the geometric characteristics of different sensors. Wang et al. (2024) introduced a fast registration method based on gray-level normalized mutual information, which effectively addresses low information content and high overlap issues through a block-based strategy, enabling fast and automatic registration. However, there is still room for improvement in matching high-resolution images with significant local differences. Exploring the combination of other mutual information metrics may further improve registration accuracy and efficiency.

## 4. Datasets

### 4.1 Optical-SAR Image Registration Datasets

This section summarizes several major open-source datasets in the field of SAR-optical image registration. Below is a detailed list of key information and download links for these datasets. Among these datasets, some contain tens of thousands of multi-modal image pairs, suitable for model training in deep learning-related research (such as registration, fusion, and recognition). Others contain tens to hundreds of image pairs from different domains or scenes, primarily for algorithm testing and evaluation. Despite significant differences in images obtained from different SAR sensors, no datasets currently contain SAR data from more than three satellites. Additionally, publicly available datasets do not provide accurate registration ground truth. To address this, our dataset uses automated registration and manual visual inspection, leveraging SAR data from four satellites, to build a comprehensive SAR-optical image registration performance evaluation dataset. This will provide a fair comparison platform for traditional algorithms and be used to assess the generalization capability of deep learning-based registration networks.

1) SARptical Dataset[1] (Wang and Zhu, 2018) contains 10,108 pairs of SAR and optical images of Berlin, Germany, captured between 2009 and 2013 by the TerraSAR-X and UltraCAM sensors. The optical images have a resolution of 0.2 meters, while the SAR images have a resolution of 1 meter, and both image types are of size 112×112 pixels.

2) SEN1-2 Dataset[2] (Schmitt et al., 2018) was captured by the Sentinel series satellites in 2017, covering seasonal surface regions across all five continents. The dataset contains a total of 282,384 pairs of SAR and optical images with a resolution of 10 meters and image size of 256×256 pixels.

3) SEN12MS Dataset[3] (Schmitt et al., 2019) contains 180,662 pairs of images, combining Sentinel-1 dual-polarized SAR data (VV and VH polarization), Sentinel-2 multispectral images, and land cover maps generated by MODIS. It covers global land areas and four meteorological seasons, with a resolution of 10 meters and image size of 256×256 pixels.

---

[1] https://syncandshare.lrz.de/getlink/fiGixjRV9idETzPgG689dGB/SARptical_data.zip

[2] https://mediatum.ub.tum.de/1436631

[3] https://mediatum.ub.tum.de/1474000

4) WHU-SEN-City Dataset[4] (Wang et al., 2019) is a SAR and optical image dataset covering 32 cities in China, combining Sentinel-1 and Sentinel-2 images. It includes various surface types such as urban buildings, farmland, forests, rivers, and lakes. The dataset has undergone rigorous preprocessing, including re-projection, cropping, and color adjustment, to ensure quality and consistency. The dataset contains 18,542 pairs of images, with a resolution of 10 meters and an image size of 256×256 pixels.

5) OS-dataset[5] (Xiang et al., 2020) contains 10,692 pairs of high-resolution optical and SAR images, with an image size of 256×256 pixels, covering multiple cities and surrounding areas worldwide. The SAR images are from the Gaofen-3 satellite with a resolution of 1 meter, while the optical images are sourced from Google Earth and resampled to a 1-meter resolution.

6) QXS-SAROPT Dataset[6] (Huang et al., 2021) contains 20,000 pairs of high-resolution optical and SAR images, with an image size of 256×256 pixels. The SAR images are from the Gaofen-3 SAR satellite, and the optical images are sourced from Google Earth, both with a resolution of 1 meter. The dataset covers three port cities: Santiago, Shanghai, and Qingdao.

7) SOPatch Dataset[7] (Xu et al., 2023) is created based on the WHU-SEN-City, OS-dataset, and SEN1-2 datasets, with further precise registration. It contains over 650,000 pairs of optical and SAR images, covering a variety of scenes such as mountains, lakes, buildings, farmland, and barren land, making it suitable for training models with good generalization ability. The SAR images are from the Gaofen-3 SAR satellite and Sentinel-1 satellite, with resolutions of 1 meter and 10 meters. The optical images are from Sentinel-2 with a resolution of 10 meters, and the image size is 64×64 pixels.

8) OSEval Dataset[8] (Xiang et al., 2023) contains 1,232 pairs of sub-meter resolution optical and SAR images, used to evaluate registration algorithms. The images are sourced from the GF-3, WorldView-2, WorldView-3, and SuperView-1 satellites. The dataset covers areas such as airports, residential areas, industrial zones, and dense urban regions, with corner points in the optical and SAR images used as ground truth. This dataset provides a fair evaluation platform for both traditional algorithms and deep learning networks.

**4.2 Introduction of Our Dataset**

To promote the development of more general multi-modal image registration methods, the MultiResSAR dataset has been constructed as a large-scale, multi-source, multi-resolution, and multi-scene dataset specifically designed for optical and SAR image registration. As shown in Fig. 3, the dataset contains 10,850 pairs of optical and SAR images, covering SAR data from four commercial satellites, with resolutions ranging from 0.16 meters to 10 meters. The scene types include urban, rural, plains, hills, mountains, and water areas. Notably, high-resolution SAR images with a 16 cm resolution provided by the Umbra satellite were selected to create a test dataset consisting of 850 pairs of ultra-high-resolution SAR images. Corresponding optical image pairs were obtained through georeferencing, as shown in Fig. 4. A large number of experimental results show that existing registration algorithms cannot consistently maintain excellent performance across different sources, resolutions, and scenes. In particular, for ultra-high-resolution SAR image tests, no algorithm has efficiently handled this type of registration task. The registration problem of high-resolution SAR and optical images has become a major challenge in the multi-modal image registration field. Current algorithms exhibit significant instability when dealing with different imaging conditions and resolution variations, especially in the registration of high-resolution SAR and optical images. The construction of the MultiResSAR dataset aims to provide a standardized data foundation to promote the generalization and robustness of registration algorithms in multi-modal and multi-resolution image registration. Through further research and optimization, we aim to develop efficient registration methods suitable for multi-source, multi-resolution, and multi-scene scenarios, thus providing stronger support for remote sensing image fusion and analysis.

---

[4] https://github.com/whu-csl/WHU-SEN-City

[5] https://pan.baidu.com/s/14bqaJhMSZEy7EXcXVAc77w?pwd=vriw

[6] https://github.com/yaoxu008/QXS-SAROPT

[7] https://pan.baidu.com/s/12yIheGOg6JTTsYAQfX7Pfg?pwd=1234

[8] https://github.com/xym2009/OS-Eval

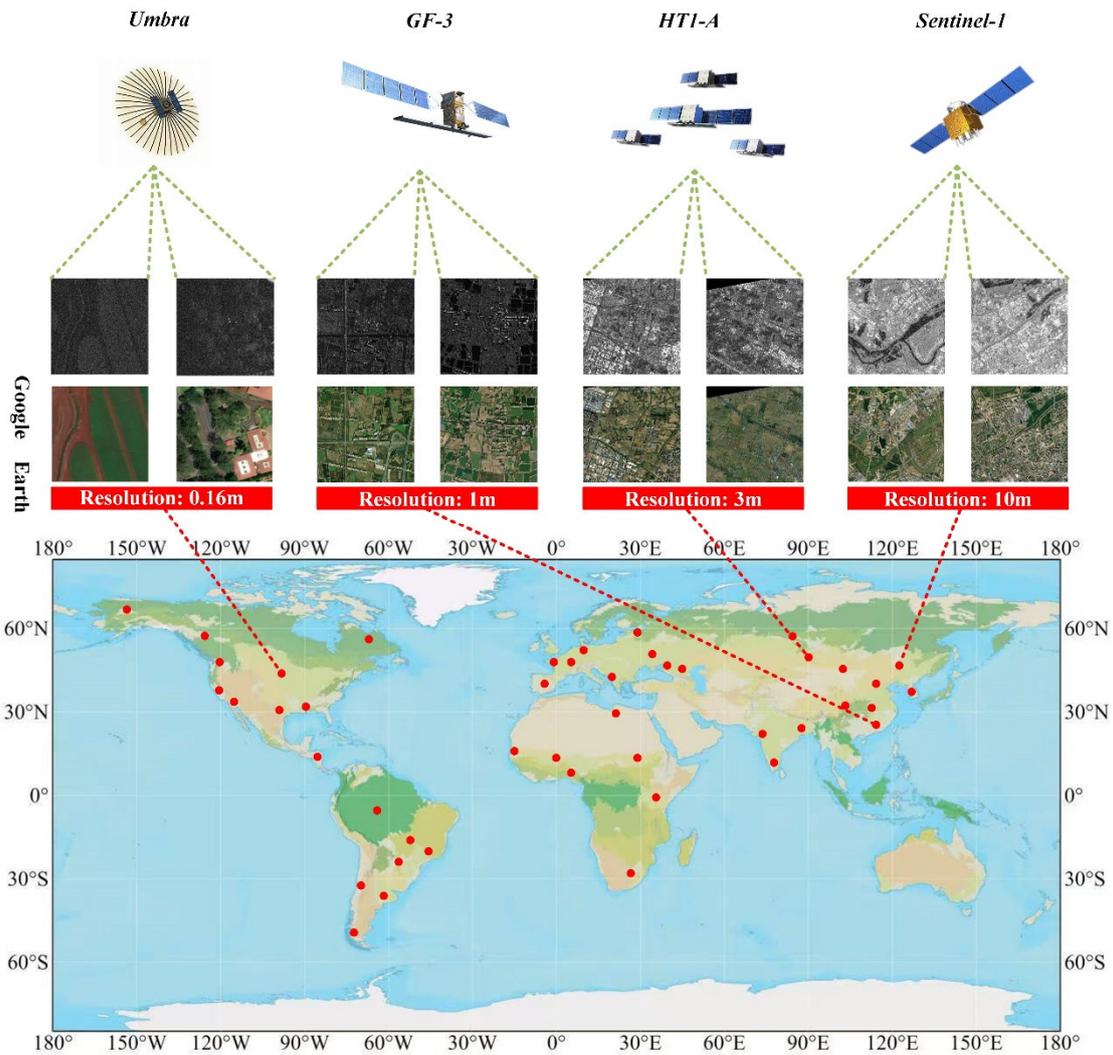

Fig. 3 Global Distribution Map of the MultiResSAR Dataset

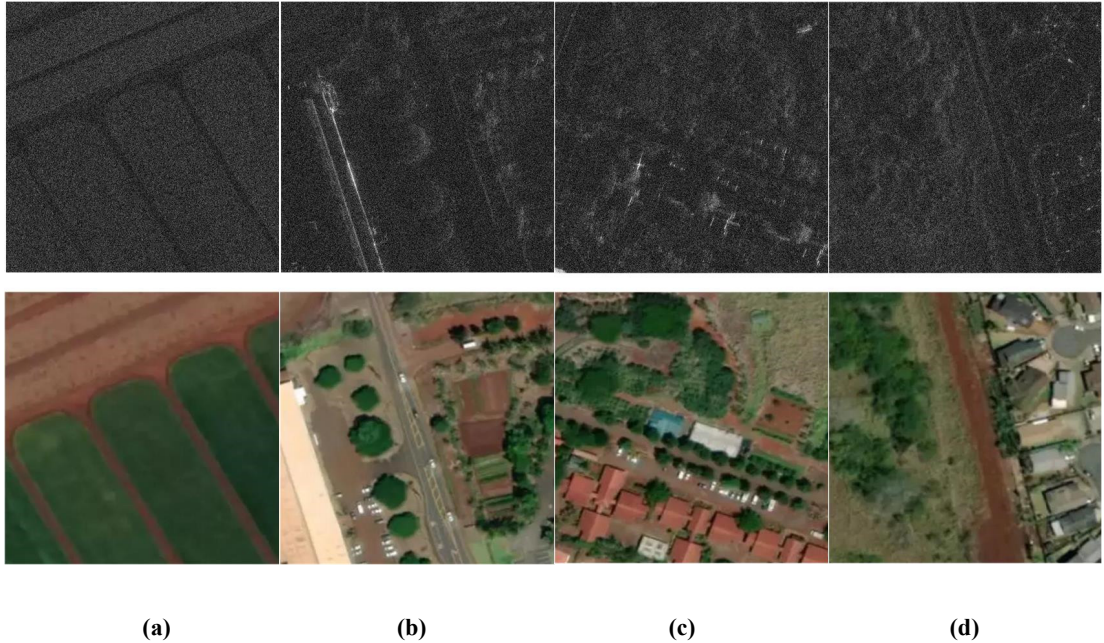

|      (a)      |      (b)      |      (c)      |      (d)      |

Fig. 4 Partial Display of the Ultra-High-Resolution SAR-Optical Image Dataset

**4.3 Ground Truth Acquisition for Dataset Registration**

Due to the fact that the original optical and SAR images come from different data sources, geometric transformations may exist between preprocessed images of the same area, making image registration a critical process. Although existing advanced algorithms offer the advantages of fast speed and high accuracy in automatic registration, their performance may be insufficient in regions with severe noise interference. Furthermore, algorithmic registration may introduce errors due to the inherent characteristics of the algorithm itself. To ensure the accuracy of the registration and reduce errors introduced by the algorithm, a combination of registration algorithms and manual verification is employed. Specifically, for each pair of optical and SAR images, based on the automatic registration results, professionals in the fields of surveying and remote sensing are invited to uniformly select several control points that are geometrically similar and least affected by noise, such as corners or other prominent features, and ensure that the error is controlled within one pixel. The specific production process is shown in the Fig. 5.

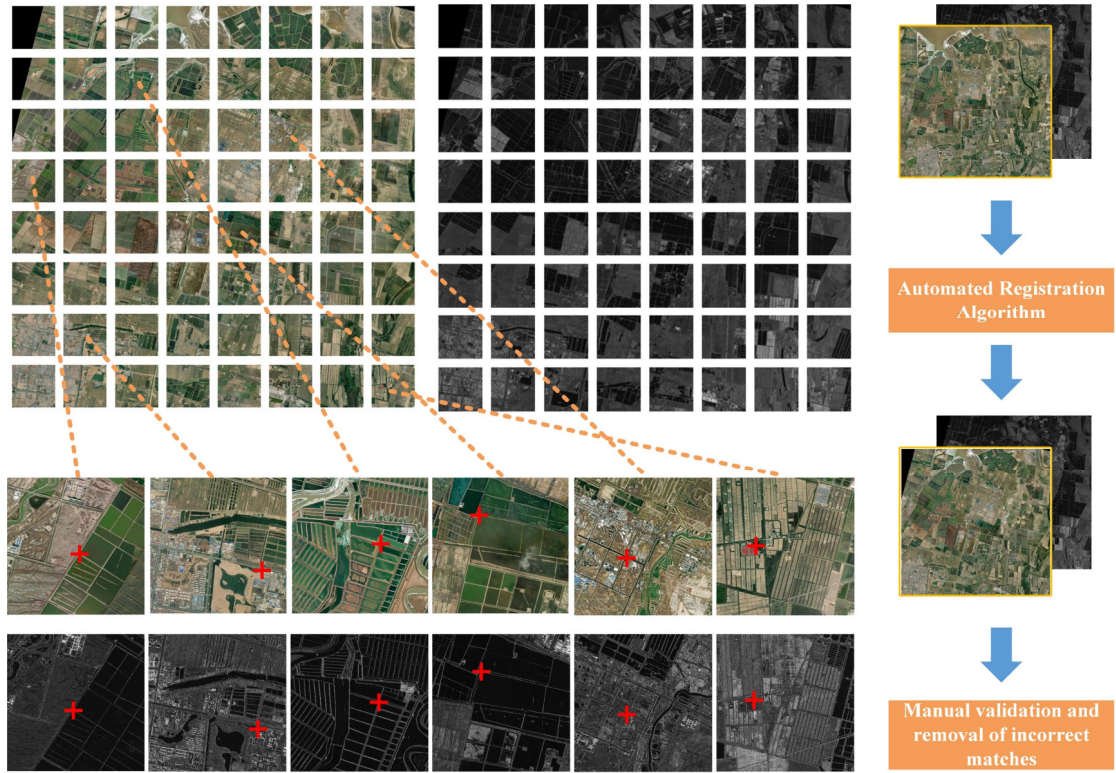

Fig. 5 Full Workflow of Image Dataset Creation and Partial Display of Block-based Registration Control Points

### 4.4 Dataset Parameters

The statistical analysis of the MultiResSAR dataset is shown in Table 5. This dataset contains SAR data from multiple sources, including Sentinel-1, GF-3, HT1-A, and Umbra, with various polarization modes, covering multiple countries and cities, and spatial resolutions ranging from 0.16 to 10 meters. The dataset includes six types of scenes from regions around the world, including urban, rural, plains, hills, mountains, and water areas. The proportions of data from Sentinel-1, GF-3, and HT1-A sources are approximately equal.

Table 5 MultiResSAR Dataset Statistics.　　**SR:** Spatial Resolution　　**GE:** Google Earth

| SAR | Optical | SR(m) | Num | Image scenes | Location |
|---|---|---|---|---|---|
| Sentinel-1 | GE | 10 | 3889 | urban, rural, plain, hills, mountain, water | China, New York, Los Angeles, Tokyo, Singapore… |
| HT1-A | GE | 3 | 3011 | urban, rural, plain, hills, mountain | China |
| GF-3 | GE | 1 | 3100 | urban, rural, plain, hills, mountain | guangzhou in China |
| Umbra | GE | 0.16 | 850 | urban, rural, plain, mountain | America |

### 5.Experiments and Discussion

In this section, an in-depth analysis and comparison of the performance of 16 algorithms on the MultiResSAR dataset is provided. These methods span a range of image registration techniques, from local feature-based approaches to deep learning methods, demonstrating their performance across different scenes and environments. This analysis offers important technical support and reference for image registration tasks. The statistical results of these 16 algorithms and their code addresses are shown in Table 6.

Transformer-based matching methods have demonstrated significant advantages in recent years. LoFTR, as a representative of these methods, leverages the global perception capability of Transformers to effectively address the challenge of matching in low-texture regions. By combining coarse-grained matching with fine-grained processing, LoFTR avoids errors commonly found in traditional stage-by-stage matching. Efficient LoFTR further optimizes this approach by adopting an aggregated attention mechanism and dual-stage correlation layers, significantly improving matching efficiency, making it more suitable for large-scale data processing and latency-sensitive applications. RoMa, on the other hand, combines pre-trained features with convolutional networks to build a feature pyramid, enhancing the robustness and accuracy of the matching process. XoFTR addresses complex matching issues related to viewpoint, scale, and texture by using cross-modal pretraining and sub-pixel refinement matching, advancing the development of image matching technologies.

Lightweight Convolutional Neural Network (CNN) methods are also widely applied in the image matching field. XFeat optimizes the CNN design by reducing the number of channels while retaining high-resolution image inputs, providing a lightweight and efficient matching solution. LightGlue, through optimization of SuperGlue, enhances matching efficiency and accuracy, with its adaptive inference capability allowing it to adapt to various scenarios. SuperGlue combines graph neural networks and attention mechanisms, improving feature matching accuracy through the learning of 3D geometric transformations, making it one of the core technologies for image registration. SGM-NET, through seed graph matching and graph neural networks, achieves efficient information transfer and matching allocation optimization, demonstrating excellent performance in both accuracy and efficiency.

Feature self-similarity methods have demonstrated excellent performance in multi-modal remote sensing imagery. MOSS utilizes multi-dimensional oriented self-similarity features for coarse-to-fine matching, making it particularly suitable for complex multi-modal remote sensing scenarios. ASS extracts self-similarity features through mean-shift filtering, constructing minimal self-similarity and index maps, which significantly enhances the algorithm's robustness to radiometric differences and noise. Matching methods based on scale invariance and phase consistency, such as SAR-SIFT and RIFT, also perform excellently. SAR-SIFT improves the classic SIFT algorithm, enhancing the stability of keypoint detection, particularly in speckle noise environments. RIFT achieves rotational invariance through phase consistency detection, improving performance in environments with nonlinear radiometric differences.

Local feature transformation and spatial domain enhancement methods, such as LNIFT and SRIF, significantly improve the accuracy of multi-modal image registration by reducing radiometric distortions and enhancing image structural information. LNIFT improves the adaptability to radiometric differences by applying local normalization feature transformations, while SRIF enhances feature matching accuracy by simplifying the scale space and performing local intensity binarization.

Finally, the DKM method, based on kernel regression and dense confidence maps, achieves precise matching in geometric estimation through global matching and distortion correction. This method utilizes dense confidence maps, effectively improving the accuracy and robustness of the matching process.

The experiments in this paper were conducted in the Matlab R2018a environment, with the following test platform configuration: Processor: AMD Ryzen 9 5900HX with Radeon Graphics, base frequency of 3.30 GHz, 64 GB of RAM, operating system: Windows 11 x64, and a deep learning platform equipped with an RTX 4090 GPU. To ensure fairness in the comparison, the experiments used the code provided by the authors, with the recommended parameter settings. The code for all 16 representative image registration methods has been made publicly available.

**Table 6** Registration Method Statistics and Code Addresses

| Algorithm | Author | Year | Code Addresses |
|---|---|---|---|
| SAR-SIFT | Dellinger et al. | 2015 | https://github.com/yishiliuhuasheng/sar_sift |
| SuperGlue | Sarlin et al. | 2020 | https://github.com/magicleap/SuperGluePretrainedNetwork |
| RIFT | Li et al. | 2020 | https://github.com/LJY-RS/RIFT-multimodal-image-matching |
| SGM-NET | Chen et al. | 2021 | https://github.com/vdvchen/SGMNet |
| LoFTR | Sun et al. | 2021 | https://github.com/zju3dv/LoFTR |
| LNIFT | Li et al. | 2022 | https://github.com/arunsahu159/LNIFT-Locally-Normalized-Image-for-Rotation-Invariant-Multimodal-Feature-Matching |
| ASS | Xiong et al. | 2022 | https://gitee.com/xxin08/ASS |
| Lightglue | Lindenberger et al. | 2023 | https://github.com/cvg/LightGlue |
| DKM | Edstedt et al. | 2023 | https://github.com/Parskatt/dkm |
| SRIF | Li et al. | 2023 | https://github.com/LJY-RS/SRIF |
| HOWP | Zhang et al. | 2023 | https://skyearth.org/publication/project/HOWP/ |
| XoFTR | Tuzcuoğlu et al. | 2024 | https://github.com/OnderT/XoFTR |
| XFeat | Potje et al. | 2024 | https://github.com/brukg/xfeat |
| RoMa | Edstedt et al. | 2024 | https://github.com/Parskatt/RoMa |
| Efficient LoFTR | Wang et al. | 2024 | https://github.com/zju3dv/efficientloftr |
| MOSS | Zhang et al. | 2024 | https://github.com/betterlll/MOSS_data |

To evaluate the performance of registration methods, researchers have developed various evaluation metrics, which can also serve as a reference for selecting registration techniques in practical applications. Evaluation methods can be divided into qualitative and quantitative assessments. Qualitative evaluation focuses on the subjective analysis of the registration results, while quantitative evaluation is not influenced by the observer's subjectivity and provides an objective and precise measure of the registration method's performance. Since each evaluation metric measures the performance of registration from a specific perspective, it is necessary to consider the performance of different metrics comprehensively during the decision-making process. In this paper, four quantitative metrics were used to measure the registration performance of the algorithms: Success Rate (SR), Number of Correct Matches (NCM), Root Mean Square Error (RMSE), and Time of Matching (TM).

1. SR refers to the ratio of the number of successfully matched image pairs to the total number of image pairs. This metric reflects the robustness of the matching method for a specific type of multi-modal image pair.

$$I(p_i) = \begin{cases} 1, & NCM(p_i) \geqslant N_{\min} \\ 0, & \text{else} \end{cases}, SR = \frac{1}{M} \cdot \sum_i I(p_i) \cdot 100\% \qquad (1)$$

$I(p_i)$ represents a logical value, 1 represents a successful matching attempt, while 0 represents a failed matching attempt. $N_{\min}$ represents the minimum number of correct matching points (set $N_{\min}$ to 20). SR represents the success rate of matching, and $M$ represents the total number of image pairs in the dataset.

2. NCM refers to the number of image pairs with more than 20 matched corresponding points, while excluding image pairs with a root mean square error (RMSE) greater than 10 pixels, serving as the count of correct matches.

3. RMSE reflects the accuracy of the matching process. The smaller the RMSE value, the higher the matching accuracy. The mathematical expression for RMSE is given by Equation 2:

$$RMSE = \sqrt{\frac{1}{N}\left(\sum_{i=1}^{N}\left[\left(x_i^{'} - x_i^{''}\right)^2 + \left(y_i^{'} - y_i^{''}\right)^2\right]\right)} \qquad (2)$$

In the equation, $N$ represents the number of ground truth points, $(x_i^{''}, y_i^{''})$ represents the coordinates of the i-th ground truth point $(x_i^{'}, y_i^{'})$ after the corresponding matching transformation.

4. TM as a metric for evaluating matching quality, focuses primarily on the time expenditure and efficiency of the matching process. It refers to the total time required from the start of matching with the input data to the output of the matching results. The shorter the matching time, the higher the algorithm's efficiency and processing speed.

Fig 6 shows the performance of 16 methods on the MultiResSAR dataset. Experimental analysis indicates that there are significant differences among these advanced matching algorithms in terms of matching accuracy, efficiency, and robustness. We conducted a detailed comparison of each algorithm from the perspectives of the number of matching points, distribution characteristics, matching accuracy, robustness, and matching speed and efficiency.In terms of the number and distribution of matching points, the LNIFT algorithm performs poorly, failing to obtain a rich set of homologous point pairs. The ASS and RIFT algorithms have dense matching points in detail-rich areas such as urban buildings and farmland, indicating strong feature detection capabilities in fine details, but they perform poorly in low-texture regions. In contrast, LoFTR and Efficient LoFTR have fewer matching points and more mismatches in scenes with significant image content variation, making them more suitable for fast matching applications rather than precise feature extraction tasks.Regarding matching accuracy, SuperGlue and LightGlue performed excellently, maintaining consistent matching results across different scenes with even point distribution and minimal mismatches, making them suitable for high-precision matching scenarios. Additionally, HOWP demonstrated good matching accuracy in texture-complex areas, stably recognizing feature points, while XFeat, DKM, and SGM-NET were more prone to mismatches in weak-texture areas.In terms of robustness, ASS and HOWP showed strong robustness in SAR and optical image matching, being able to identify prominent feature points. In contrast, RoMa and DKM had more mismatches in scenes with significant feature changes, indicating insufficient robustness to different image features.Regarding matching speed and efficiency, Efficient LoFTR demonstrated fast matching speed, suitable for real-time applications. Although it compromises on matching accuracy, it excels in real-time processing tasks. SuperGlue and LightGlue have longer computation times but deliver superior matching results, making them suitable for tasks with high accuracy requirements and offline processing allowances. RIFT and XFeat strike a good balance between speed and accuracy, making them suitable for scenarios requiring both real-time processing and higher precision.In summary, SuperGlue and LightGlue's high-precision performance is ideal for accuracy-demanding scenarios, while Efficient LoFTR excels in real-time tasks. RIFT and RoMa perform stably across various scenarios, making them a preferred option for balancing accuracy and efficiency. However, in regions with significant terrain changes or weak textures, no algorithm has been able to maintain consistently outstanding matching performance.

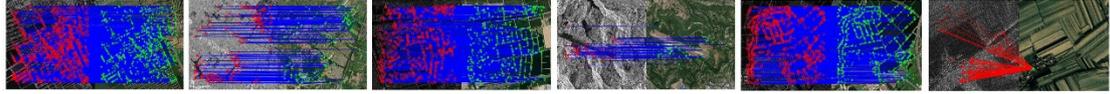

(**a**) RIFT

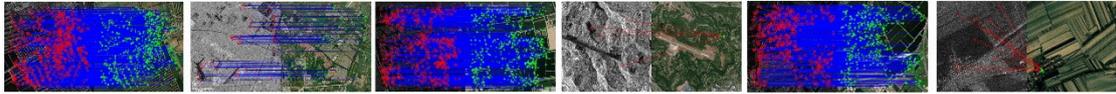

(**b**) HOWP

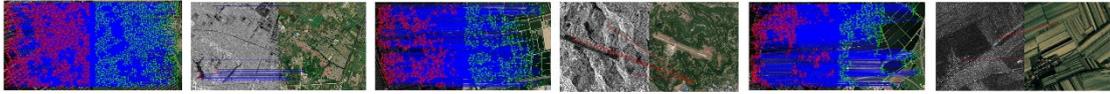

(**c**) XoFTR

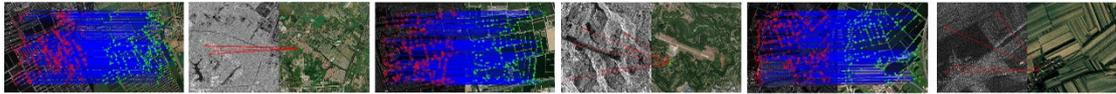

(**d**) ASS

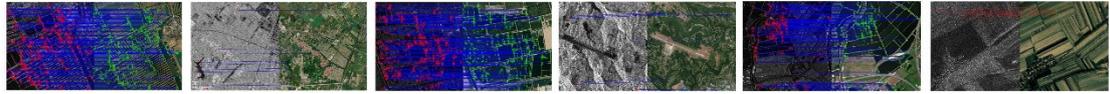

(**e**) XFeat

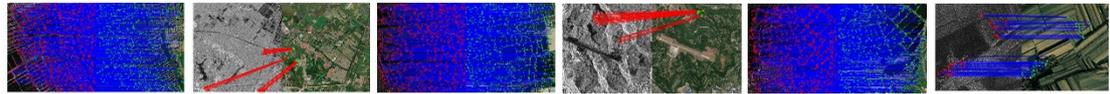

(**f**) RoMa

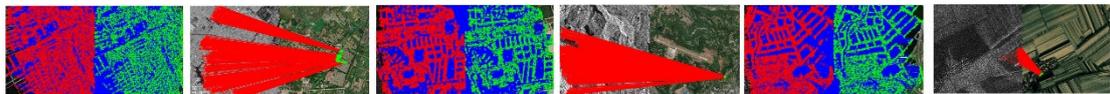

(**g**) DKM

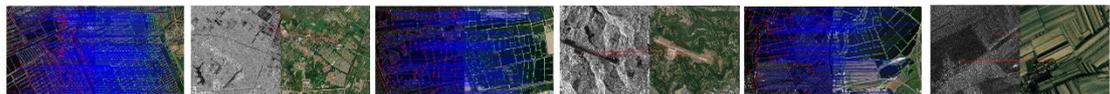

(**h**) SuperGlue

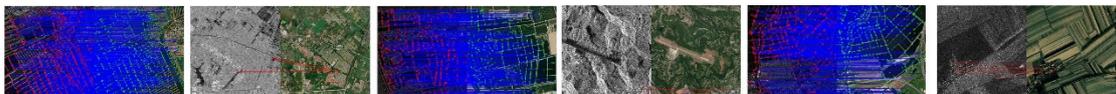

(**i**) Lightglue

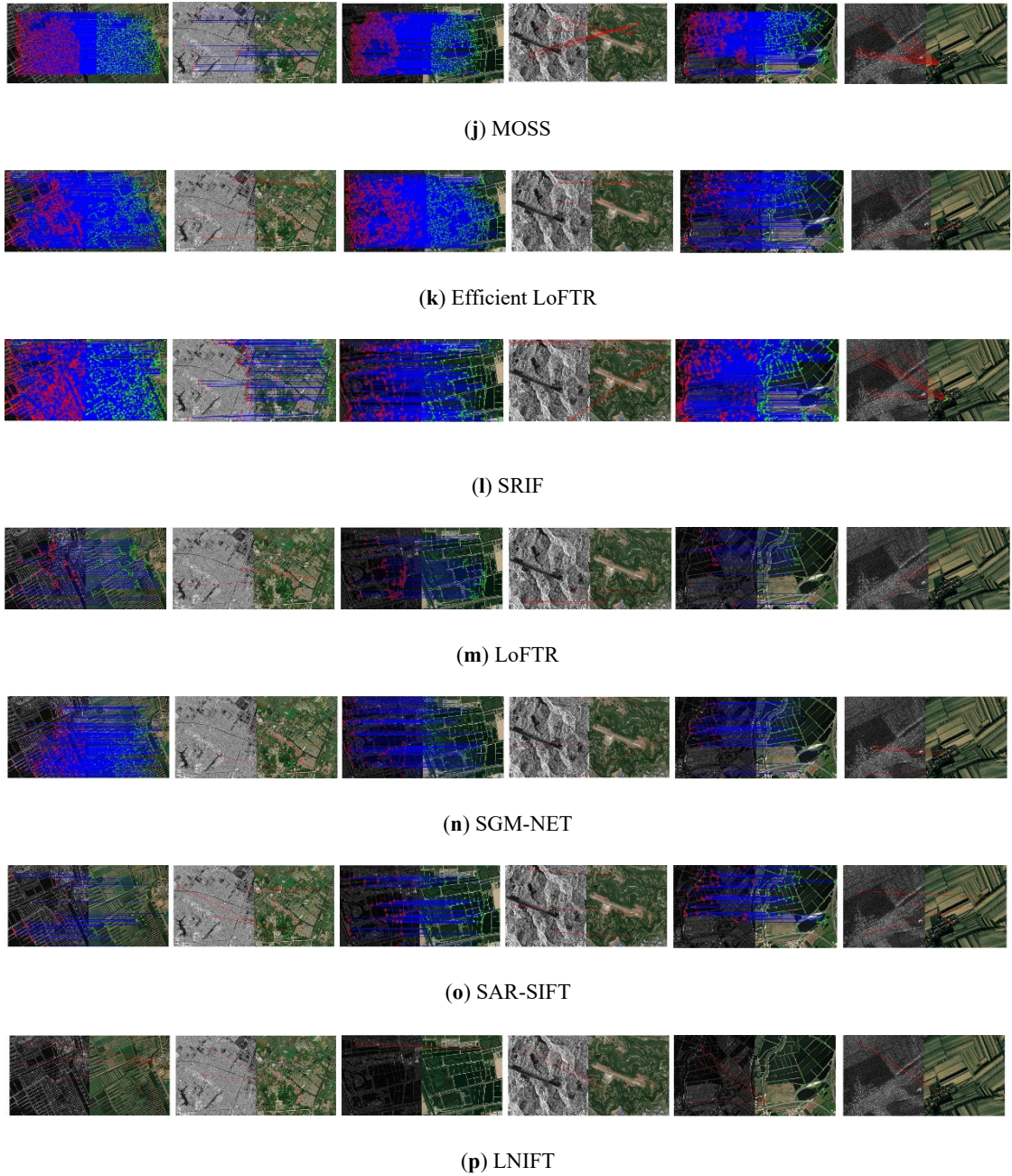

(**j**) MOSS

(**k**) Efficient LoFTR

(**l**) SRIF

(**m**) LoFTR

(**n**) SGM-NET

(**o**) SAR-SIFT

(**p**) LNIFT

Fig. 6 Matching Results of 16 Comparison Methods

An in-depth analysis and comparison of 16 advanced matching algorithms were conducted on the MultiResSAR dataset. Table 7 presents the average results for four metrics: SR (units: %), NCM (units: number of points), RMSE (units: pixels), and TM (units: seconds). Fig. 7 displays the matching success rate of each algorithm on the MultiResSAR dataset, Fig. 8 shows the RMSE of each algorithm and Fig. 9 shows the NCM for each algorithm.

**Table 7** Registration Performance Evaluation Results for the MultiResSAR Datase

|  | Algorithm | SR | Rmse | NCM | Time |
|---|---|---|---|---|---|
| Deep | XoFTR | **40.58%** | **3.03** | 244.26 | 0.032 |
| Learning | RoMa | 35.26% | 3.15 | **589.70** | 1.065 |
| Methods | XFeat | 36.29% | 4.94 | 73.24 | **0.017** |

|              | Method         |         |      |        |        |
|--------------|----------------|---------|------|--------|--------|
|              | DKM            | 29.85%  | 3.41 | 395.96 | 0.443  |
|              | SuperGlue      | 29.02%  | 3.39 | 74.03  | 1.926  |
|              | Lightglue      | 26.41%  | 3.34 | 122.85 | 0.042  |
|              | Efficient LoFTR| 15.71%  | 3.70 | 100.17 | 0.040  |
|              | LoFTR          | 9.50%   | 3.68 | 80.77  | 0.037  |
|              | SGM-NET        | 2.05%   | 3.16 | 122.15 | 0.550  |
|              | RIFT           | **66.51%** | **3.58** | 108.40 | 5.283 |
|              | HOWP           | 52.63%  | 4.11 | 115.58 | 12.420 |
| Traditional  | ASS            | 39.34%  | 3.83 | 86.31  | 5.717  |
| Methods      | MOSS           | 17.93%  | 3.98 | 114.36 | 9.666  |
|              | SRIF           | 13.28%  | 5.35 | **230.06** | 7.373 |
|              | SAR-SIFT       | 1.68%   | 4.47 | 38.59  | 3.569  |
|              | LNIFT          | 0.41%   | 4.71 | 61.53  | **1.602** |

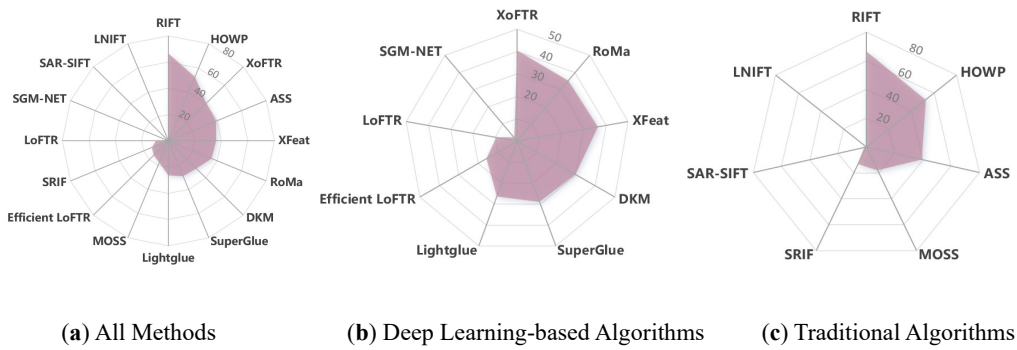

(a) All Methods  (b) Deep Learning-based Algorithms  (c) Traditional Algorithms

Fig. 7  SR results for 16 algorithms

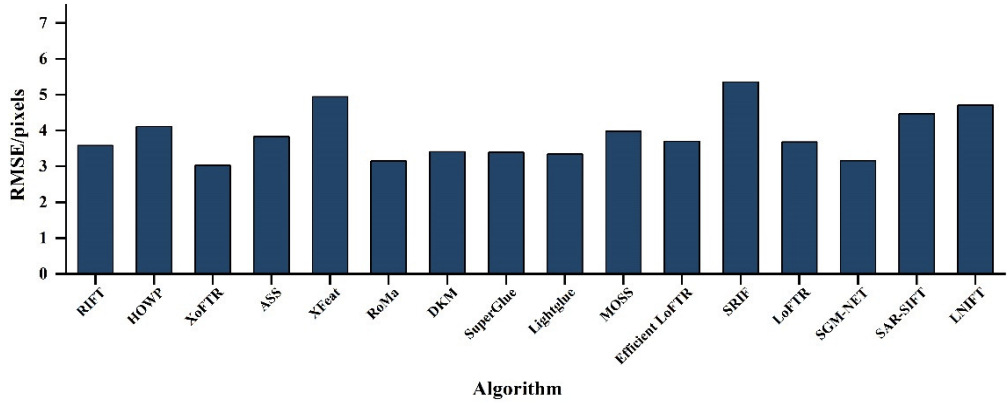

Fig. 8  RMSE results for 16 algorithms

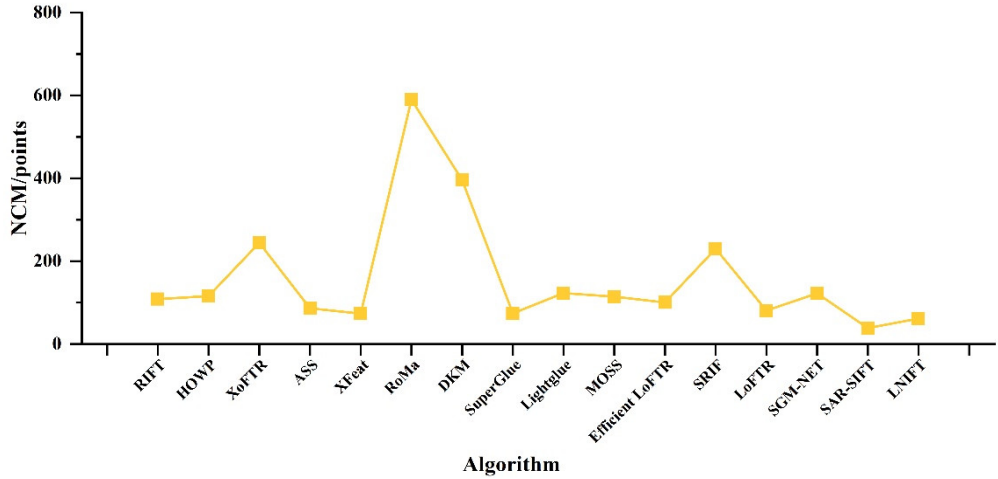

Fig. 9    NCM results for 16 algorithms

Based on the experimental results in Table 7, the NCM, SR and RMSE of each algorithm on the MultiResSAR dataset show significant differences. No algorithm currently achieves 100% successful matching across different resolutions and scenes. As the resolution of SAR and optical images increases, the registration performance gradually decreases, with almost all matches failing in sub-meter resolution image pairs. As shown in the Fig. 9 , the RoMa algorithm performs exceptionally well in the NCM metric, reaching 589.70 points, demonstrating a significant advantage in the number of correct matching points. It is followed by DKM, XoFTR, and SRIF, with NCM values of 395.96, 244.26, and 230.06, respectively. The NCM values for other algorithms range from 30 to 130 points, showing considerable variation. In terms of matching SR, among deep learning algorithms, the XoFTR algorithm currently achieves the best performance, but with a matching success rate of only 40.58%. Among traditional algorithms, RIFT performs the best with a success rate of 66.51%. Most other algorithms have a matching success rate below 50%. Notably, only RIFT, HOWP, and XoFTR achieve a success rate above 40%. It is noteworthy that RoMa achieves a high matching success rate while also obtaining a large number of NCMs. For RMSE, XoFTR has the lowest RMSE at 3.03, indicating high matching accuracy and strong stability. In contrast, SRIF and XFeat have higher RMSE values of 5.35 and 4.94, respectively, indicating lower matching accuracy. RIFT and RoMa have moderate RMSE values, showing a balance between accuracy and robustness. A large number of experiments indicate that no algorithm currently maintains outstanding performance across different data sources, resolutions, and scenes.

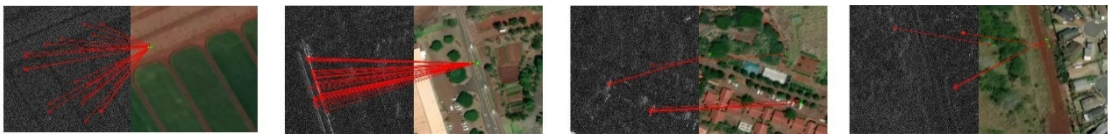

(a) RIFT

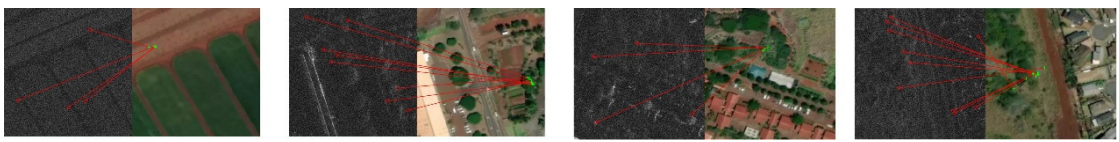

**(b)** HOWP

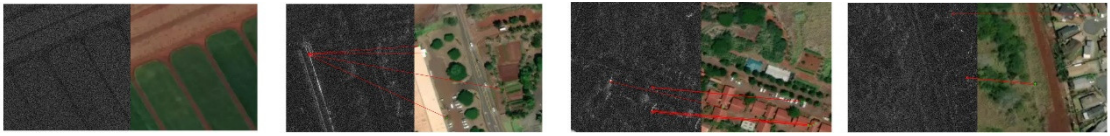

**(c)** XoFTR

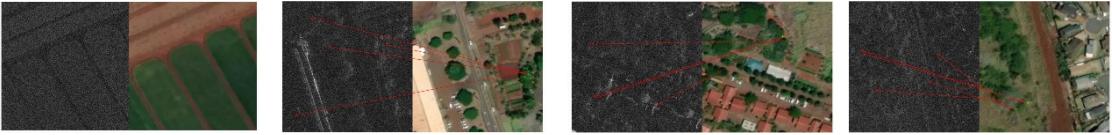

**(d)** ASS

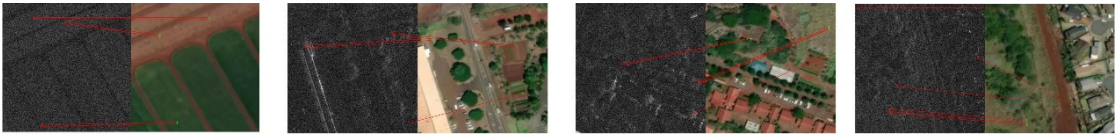

**(e)** Xfeat

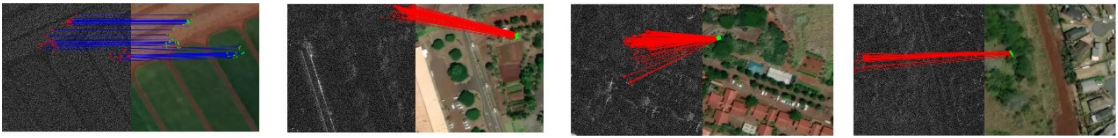

**(f)** RoMa

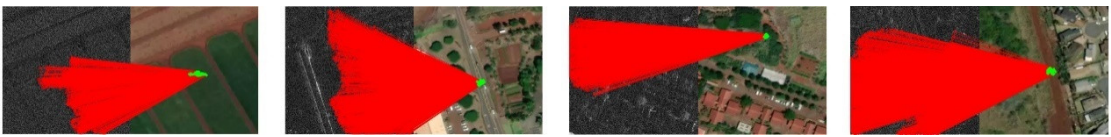

**(g)** DKM

Fig. 10 Ultra-High-Resolution Dataset Test Results

Sixteen advanced matching algorithms were tested on a dataset of 850 pairs of ultra-high-resolution SAR images, and the experimental results are shown in Fig. 10. In sub-meter resolution image pairs, nearly all matches failed, with only the RoMa algorithm achieving correct registration results in the four image groups. Due to the interference of multiplicative speckle noise in ultra-high-resolution SAR images, the number of correct matching points obtained by the RoMa algorithm is relatively small and unevenly distributed. As SAR and optical images progress toward ultra-high resolution, addressing the challenges of mitigating the interference of multiplicative speckle noise, overcoming fine registration difficulties caused by complex structures such as buildings and weak textures, and effectively extracting stable common features for ultra-high-resolution SAR-optical image registration, will be critical scientific research topics for high-precision registration and fusion.

## 6. Challenges and Issues

By reviewing the current state of area-based registration, feature-based registration, deep learning, and data-driven high-resolution SAR-optical image matching, this section organizes and analyzes the available datasets and registration evaluation metrics for optical and SAR image registration. Although existing research has largely addressed nonlinear radiometric and geometric distortions caused by modal differences, many challenges still exist in practical remote sensing applications. With the development of remote sensing technology and the increase in application scenarios, the precision and adaptability of SAR and optical image registration urgently need further improvement. The main issues faced and future research directions are as follows:

**1) Lack of Large-Scale, Multi-Source, Multi-Resolution, and Multi-Scene Registration Datasets**. Currently, there is a shortage of fusion datasets for SAR and optical images, particularly large-scale datasets with multi-source, multi-resolution, and multi-scene data. This limits the generalization ability of deep learning-based registration methods and hinders their adaptability and robustness in practical applications. Future research should focus on the development and sharing of datasets to meet the demands for multi-scene and multi-resolution registration. On one hand, more SAR images can be collected through open-source satellite platforms to generate multi-scene, multi-resolution datasets for registration research. On the other hand, unsupervised learning and few-shot learning offer potential solutions to the data scarcity problem, helping achieve better registration performance with limited datasets.

**2) Challenges in Feature Detection and High-Precision Registration for Ultra-High-Resolution Imagery**. As image resolution increases, the multiplicative speckle noise and structural features in SAR images become deeply intertwined, making the detection of common features and dense corresponding point recognition more challenging. Additionally, terrain undulations and three-dimensional structures (such as buildings and mountains) have a significant impact on image registration, particularly in complex scenes such as urban areas, where they introduce noticeable geometric distortions. Future research should focus on suppressing SAR noise, improving the efficiency of utilizing three-dimensional image information, developing high-precision geometric registration models, and enhancing feature matching and stability. These advancements will support high-precision registration and corresponding point identification.

**3) Cross-View SAR-Optical Image Registration**. The differences in observation angles of the same area from different perspectives present geometric consistency challenges for wide-baseline cross-view SAR and optical image registration, leading to lower registration accuracy. Future research could explore optimization methods to address these viewpoint differences, including the development of viewpoint transformation models and improvements in feature matching strategies, to enhance the registration performance of cross-view images.

**4) In areas with weak texture, repetitive texture, and significant intensity variation, as well as in cases where images exhibit severe radiometric distortions and complex geometric deformations.** Traditional registration methods often struggle to achieve optimal results due to insufficient local features, sparse distribution, and the complexity of radiometric differences and deformations. Future research could focus on designing more robust registration algorithms for these complex scenarios, using adaptive feature extraction and matching strategies to address issues such as weak textures, repetitive textures, and intense intensity changes. Additionally, combining local appearance and structural features under multi-modal conditions can help resolve the registration challenges posed by radiometric distortions and geometric deformations. Furthermore, optimizing feature detection and description operators, exploring

acceleration algorithms such as Fast Fourier Transform (FFT), and improving outlier removal strategies are expected to significantly enhance registration accuracy and reliability while improving computational efficiency.

## 7. Conclusion

In recent years, the registration of SAR and optical remote sensing images has played a crucial role in key applications in the field of remote sensing, including image fusion, change detection, target recognition, and image stitching. With the continuous advancement of remote sensing technology, significant progress has been made in SAR and optical image registration research over the past few decades, successfully addressing issues such as nonlinear radiometric differences and geometric distortions caused by modal differences, significantly improving registration accuracy and efficiency. However, the significant differences between SAR and optical images in terms of imaging mechanisms, geometric distortions, and radiometric characteristics, especially in high-resolution SAR and optical images, still pose a severe challenge to the stability and accuracy of registration.

This paper systematically reviews the existing SAR and optical image registration methods, which can be broadly categorized into area-based, feature-based, and deep learning-based approaches. Through a brief overview of the core methods and theoretical frameworks, it provides a solid theoretical foundation and reference for further improving registration performance. However, there are two main shortcomings in the current research: first, the lack of publicly available registration datasets with multiple resolutions and scenes; and second, the lack of a systematic review and multi-level analysis of the progress in SAR and optical image registration research. To address these issues, this paper summarizes the research progress in SAR and optical image registration from the perspective of data resolution, deeply analyzes the major challenges currently faced, and explores future directions and trends.

To promote the application testing and further research of registration methods, this paper constructs and releases a dataset named MultiResSAR, which covers 10,850 pairs of multi-resolution and multi-scene SAR and optical images. The MultiResSAR dataset utilizes data from four SAR satellites, combining automatic registration and manual visual inspection to create a comprehensive SAR-optical image registration performance evaluation platform. This platform not only provides a fair comparative benchmark for traditional algorithms but also serves to evaluate the generalization capabilities of deep learning-based registration networks.

Through experimental evaluation of 16 state-of-the-art (SOTA) registration algorithms, both qualitative and quantitative assessment methods were used to visually demonstrate the performance of different methods in SAR and optical image registration. The experimental results show that: (1) Currently, no algorithm achieves 100% successful matching across different resolutions and scenes. As the resolution of SAR and optical images increases, the registration performance gradually declines, with almost all matching failing in sub-meter resolution data pairs; (2) Among deep learning algorithms, the XoFTR algorithm performs the best, achieving a matching success rate of 40.58%. Among traditional algorithms, the RIFT algorithm performs the best, with a matching success rate of 66.51%, while most other algorithms have a matching success rate below 50%; (3) Future research urgently needs to focus on noise suppression, 3D geometric information fusion, cross-view geometric transformation model construction, and deep learning model optimization to advance the stable registration of high-resolution SAR and optical images.

This paper systematically evaluates existing algorithms, revealing the bottlenecks and shortcomings of current technologies. With the rapid development of integrated sky-ground multi-sensor stereoscopic observation technologies, research focus is gradually shifting towards intelligent, precise, and real-time solutions. Although significant progress has been made in SAR and optical image registration technologies, achieving high accuracy and strong generalization under complex conditions such as multi-modality, multi-resolution, and multi-scenario still presents many challenges. Future research should focus on noise suppression, 3D geometric information fusion, cross-view geometric transformation model construction, and deep learning model optimization to promote stable registration of high-resolution SAR and optical images. SAR and optical image registration will have a profound impact in fields such as remote sensing image fusion, change detection, and intelligent target recognition, providing strong technological support for critical social applications such as land resource monitoring, disaster prevention and mitigation, and environmental protection.

## Acknowledgments

This work was supported in part by the Key Program of the National Natural Science Foundation of China under project number 42030102, the National Natural Science Foundation of China under project number 42471470 and 42401534.

## Data availability statement

The datasets generated during the current study are openly available in [Github] at https://github.com/betterlll/Multi-Resolution-SAR-dataset-.